\documentclass[twocolumn,3p,times,procedia]{elsarticle}

\usepackage{ecrc}
\usepackage{graphicx}
\usepackage{balance}
\usepackage{array,threeparttable}
\usepackage{url}
\usepackage{subcaption}

\volume{00}

\firstpage{1}

\runauth{}

\jid{procs}

\usepackage{amssymb}
\usepackage{amsmath}
\usepackage{multicol}
\usepackage{algorithm}
\usepackage{algpseudocode}

\usepackage[figuresright]{rotating}
\usepackage{bm}
\usepackage{caption}
\usepackage{titlesec}
\titleformat*{\section}{\small \bf}
\titleformat*{\subsection}{\small \em}
\titleformat*{\subsubsection}{\small \em}

\usepackage{xcolor}
\renewcommand{\textcolor}[2]{#2}

\usepackage{fancyhdr}
\fancyheadoffset[RO,EL]{0pt}
\usepackage{geometry}
\begin{document}

\begin{frontmatter}




\dochead{}
\title{
\begin{flushleft}
{\LARGE DRL-Based Federated Self-Supervised Learning for Task Offloading and Resource Allocation in ISAC-Enabled Vehicle Edge Computing} 
\end{flushleft}
}
 %

\author[]{ \leftline {Xueying Gu$^a$, Qiong Wu $^*$$^a$, Pingyi Fan$^b$, Nan Cheng$^c$, Wen Chen$^d$, Khaled B. Letaief$^e$}}

\address{ \leftline {$^a$ School of Internet of Things Engineering, Jiangnan University, Wuxi 214122, China}

  \leftline {$^b$Department of Electronic Engineering, Beijing National Research Center for Information Science and Technology, Tsinghua University, Beijing 100084, China}
	
  \leftline {$^c$State Key Lab. of ISN and School of Telecommunications Engineering, Xidian University, Xi’an 710071, China}
	
  \leftline {$^d$Department of Electronic Engineering, Shanghai Jiao Tong University, Shanghai 200240, China}
	
  \leftline {$^e$Department of Electrical and Computer Engineering, the Hong Kong University of Science and Technology (HKUST), Hong Kong 999077, China}
	

}

\cortext[]{\textcolor{red}{Corresponding author.}}

\fntext[]{Email addresses: xueyinggu@stu.jiangnan.edu.cn(X. Gu), fpy@tsinghua.edu.cn(P. Fan), dr.nan.cheng@ieee.org(N. Cheng), wenchen@sjtu.edu.cn(W. Chen), eekhaled@ust.hk(K. B. Letaief)}


\begin{abstract}

Intelligent Transportation Systems (ITS) leverage Integrated Sensing and Communications (ISAC) to enhance data exchange between vehicles and infrastructure in the Internet of Vehicles (IoV). This integration inevitably increases computing demands, risking real-time system stability. Vehicle Edge Computing (VEC) addresses this by offloading tasks to Road Side Unit (RSU), ensuring timely services. Our previous work FLSimCo algorithm, which uses local resources for federated Self-Supervised Learning (SSL), though vehicles often can't complete all iterations task. Our improved algorithm offloads partial task to RSU and optimizes energy consumption by adjusting transmission power, CPU frequency, and task assignment ratios, balancing local and RSU-based training. Meanwhile, setting an offloading threshold further prevents inefficiencies. Simulation results show that the enhanced algorithm reduces energy consumption, improves offloading efficiency and the accuracy of federated SSL.

\end{abstract}

\begin{keyword}

\textcolor{red}{Integrated sensing and communications (ISAC), Federated self-supervised learning, Resource allocation and offloading, Deep reinforcement learning (DRL), Vehicle edge computing (VEC)}


\end{keyword}

\end{frontmatter}


\section{Introduction}
As Intelligent Transportation Systems (ITS) evolve, they increasingly rely on the integration of sensing and communication technologies, collectively known as Integrated Sensing and Communications (ISAC), to significantly improve the efficiency of data exchange between vehicles and infrastructure within the Internet of Vehicles (IoV) \cite{wzhuang2019,4}. However, this also brings about significant computing demands that often exceed local computing capability, potentially hindering real-time data processing \cite{kzheng2015,rdeng2023,3}. To address these escalating computing demands, Vehicle Edge Computing (VEC) has emerged as a viable solution by deploying computing resources at the network edge, specifically at Road Side Units (RSUs) \cite{gluo2023,fwu2023,fan1}. As integral components of the vehicular communication infrastructure, RSU is equipped with computing power and storage capacity, thereby delivering real-time computing services and ensuring timely task completion \cite{ycui2020,syue,fan2}. Meanwhile, offloading partial computing task to RSU can significantly reduce the computing burden on moving vehicles in IoV.

We previously proposed a Self-Supervised Learning (SSL) method based on Federated Learning (FL) called FLSimCo algorithm \cite{FLSimCo}. In each round of FL, each vehicle can calculate the expected number of local training iterations for SSL based on utilizing the entire computing resource of vehicle. However, for efficiency and practicality, vehicle often has multiple local tasks that need to process within a fixed time, including system maintenance, running other applications, and background processes \cite{jshen2024,1,2}. This need for local task processing introduces competition for computing resources, complicating the execution of expected local training iterations. Due to CPU usage by other tasks, the expected local training iterations are often hard to fully execute in practice. In this case, offloading partial training iterations task to RSU can ensure the expected local training iterations task is completed within one round of FL \cite{zwang2017,lyang2018}. 

Compared to the extensive computing resources available in centralized cloud servers, the effective allocation of RSU resources becomes critical to meeting the demands of numerous task offloading requests from vehicles, particularly during peak periods \cite{mzhu2020,cyou2016}. A simple solution is to deploy more RSUs, but this significantly increases costs of info-structure and may lead to substantial resource waste during off-peak periods. Therefore, this paper aims to maximize the utilization of a single RSU in a four-way roundabout, rationally allocate its computing resource, and meet more vehicle offloading requests \cite{wwang2024,5}.

In this paper, we further explore the potential of efficient offloading decisions based on the FLSimCo algorithm in IoV. We propose an improved algorithm that BS employs SAC algorithm to allocate transmission powers from vehicles to BS and RSU, CPU frequencies in vehicles, assignment ratios for computing resource in RSU. According to the allocated assignment ratio, the computing resource in RSU allowed for each vehicle can be calculated. Then according to the allowed computing resource, the local training iterations task in each round of FLSimCo is divided into local and RSU assisted training. However, at lower assignment ratio, it suggests the allocated computing resource in RSU is small. In this case, if vehicle offloads task to RSU, it does not ease the computing burden on vehicle and also increases communication costs on the requested tasks. Therefore, we also set an offloading threshold to optimize offloading efficiency. After offloading partial iterations task to RSU, the vehicle processes the remaining iterations task locally. If there are still unfinished iterations in round of FL, the remaining number of iterations will be stored in local buffer of vehicle, awaiting for the next training\footnote{The source code has been released at: https://github.com/qiongwu86/Federated-SSL-task-offloading-and-resource-allocation}\footnote{\textcolor{red}{This work was supported in part by the National Natural Science Foundation of China under Grant 61701197 and Grant 62071296; in part by the National Key Research and Development Program of China under Grant 2021YFA1000500(4); in part by the National Key Project under Grant 2020YFB1807700; in part by Shanghai Kewei under Grant 22JC1404000; in part by the Research Grants Council under the Areas of Excellence Scheme under Grant AoE/E-601/22-R; and in part by the 111 Project under Grant B23008.}}.

To summarize, the primary contributions of this paper are as follows:

\begin{itemize}
	\item We propose a novel collaborative offloading method, based on the known FLSimCo algorithm, in the framework of federated SSL, which divides local iterations task into local and RSU assisted training in each round. By fully utilizing vehicle local computing resources and reasonably offloading partial task to RSU, significantly reduces the computing burden on moving vehicles in IoV.
	
	\item We use DRL algorithm for dynamic allocation of transmission power and assignment ratio to optimize system energy consumption. By adjusting transmission power and task assignment ratio, it can achieve minimizing energy consumption and improving the overall system efficiency. 
	
	\item We introduce a task offloading threshold mechanism to enable rapid response under different network conditions and computing tasks, ensuring the efficiency of task offloading. This further enhances the utilization of computing resource on both the vehicle and RSU sides in IoV, increasing the efficiency and stability of the task offloading process.
	
\end{itemize}

The remaining sections of this paper are as follows. Section \ref{sec2} provides a review of related work. Section \ref{sec3} details the system model, including the computation model, channel model, and transmission model. In Section \ref{sec4}, we formulate an optimization problem for resource allocation. In Section \ref{sec5}, we introduce the DRL-based algorithm to solve the problem proposed in previous section. In Section \ref{sec6}, we present the simulation results, demonstrating the efficacy of our algorithm. Finally, in Section \ref{sec7}, we conclude the paper by summarizing the new findings.

\section{Background and related work}
\label{sec2}
In IoV, task offloading and computing resource allocation have been longstanding research hot spots \cite{fwu2024,ycui2021,6}. With the increasing number of vehicles and the rising quantity and complexity of computing tasks, efficiently allocating and managing resource have become a focal point of attention. In this section, we will introduce the related work about task offloading and resource allocation.

\subsection{DRL-based task offloading and resource allocation}

With the rapid development of machine learning, an increasing number of work focus on applying machine learning to task offloading and resource allocation \cite{ncheng2019,rsun2024}. Due to its superior performance in handling complex dynamic environments, DRL has gradually become a popular choice for optimizing task offloading and resource allocation. 

Many work employed DRL methods for task allocation and resource allocation. In \cite{jxu2021}, Xu \emph{et al.} proposed a mobile-compatible offloading and resource allocation scheme based on the DRL method (i.e., Deep Q-Network (DQN)) aiming to minimize system cost. In \cite{jyu2022}, Yu \emph{et al.} proposed an optimization strategy for task offloading and resource allocation for static devices based on the DRL method (i.e., Deep Deterministic Policy Gradient (DDPG)), aiming to minimize system energy consumption and resource overhead. These methods are primarily designed for scenarios with limited mobility. In high-mobility scenarios, the aforementioned methods are not applicable because they do not consider the real-time changes of the target position and environment. 

In \cite{schen2020}, Chen \emph{et al.} proposed an intelligent task offloading algorithm for Unmanned Aerial Vehicle (UAV) edge computing networks aimed at minimizing latency. The algorithm offloads all training tasks to edge servers and utilizes Deep Neural Networks (DNNs) for resource allocation. In \cite{bwang2023}, Wang \emph{et al.} utilized an improved DRL method (i.e., DQN) to decide whether to execute tasks locally on vehicles or offload them to RSU in IoV. The goal is to achieve the lowest transmission and computation latency for both vehicles and RSU. In \cite{hcui2022}, Wang \emph{et al.} proposed an offloading and resource allocation scheme in IoV based on the DRL method (i.e., Advantage Actor-Critic (A2C)), aiming to minimize latency and energy consumption.

Additionally, to improve offloading efficiency, some work has started using threshold to determine whether to offload. In \cite{wchen2016}, Chen \emph{et al.} proposed that when the BS's offloading capacity is limited, setting an offloading threshold can significantly reduce signaling overhead. In \cite{oai2023}, Al-Tuhafi and Al-Hemiary proposed an algorithm that dynamically adjusts the offloading threshold based on load increase or decrease to minimize response time. In \cite{xqin2022}, Qin \emph{et al.} proposed a distributed threshold-based offloading algorithm. When the number of tasks on a device reaches the threshold, the tasks are offloaded to the cloud server for processing; otherwise, they are processed locally.

\subsection{\textcolor{red}{Task offloading in federated SSL}	}

Some work combined FL and SSL \cite{njahan2023,mfeng2022}, but these methods mostly based on static environments, relying heavily on extensive negative samples during the SSL process, which leads to extended training duration. However, in real-time demanding environment such as IoV, constraints related to time and environmental variability become significant. Therefore, conducting SSL in IoV necessitates addressing the challenge of reducing dependency on negative samples during the training process.

In \cite{swei2023}, Wei \emph{et al.} combined FL and SSL method (i.e.,  Momentum Contrast (MoCo)), and  employed dictionary to reduce the demand for a large number of negative samples during training. However, this method requires sending each vehicle's small dictionary to the BS to form a large dictionary, which undoubtedly increases communication costs. In previous work, we proposed FLSimCo algorithm, a federated SSL algorithm without employing dictionary. However, in this algorithm, we also assume each vehicle can complete all local iterations task assigned by BS in each round. To enhance efficiency, vehicles need to handle multiple tasks per round, making it difficult to complete all iteration tasks in a fixed time. Offloading partial task to RSU can address this issue.

Meanwhile, some work has been done on DRL-based task offloading within the FL framework. In \cite{swu2024}, Wu \emph{et al.}, in the context of Industrial Internet of Things (IIoT), combined FL and DRL method to complete offloading task, aiming to improve offloading success rates. In \cite{lzang2022}, Zeng \emph{et al.} proposed an online task offloading and resource allocation algorithm based on Federated DRL (i.e., DQN) in IoT, with the aim to regulate convergence speed, execution delay, overall computation rate and stability. In \cite{tzhao2023}, Zhao \emph{et al.} proposed an algorithm based on the FL and DRL method (i.e., Twin Delayed Deep Deterministic Policy Gradient (TD3)) in IIoT with Mobile Edge Computing (MEC). The aim is to minimize latency and energy consumption while maximizing security rates.

However, to the best of our knowledge, there is no existing work that utilizes DRL for threshold-based task offloading and resource allocation within the federated SSL framework.

\section{System model}
\label{sec3}
\begin{figure}
	\centering
	\includegraphics[scale=0.3, trim=0.1cm 0.1cm 0.1cm 0.1cm, clip]{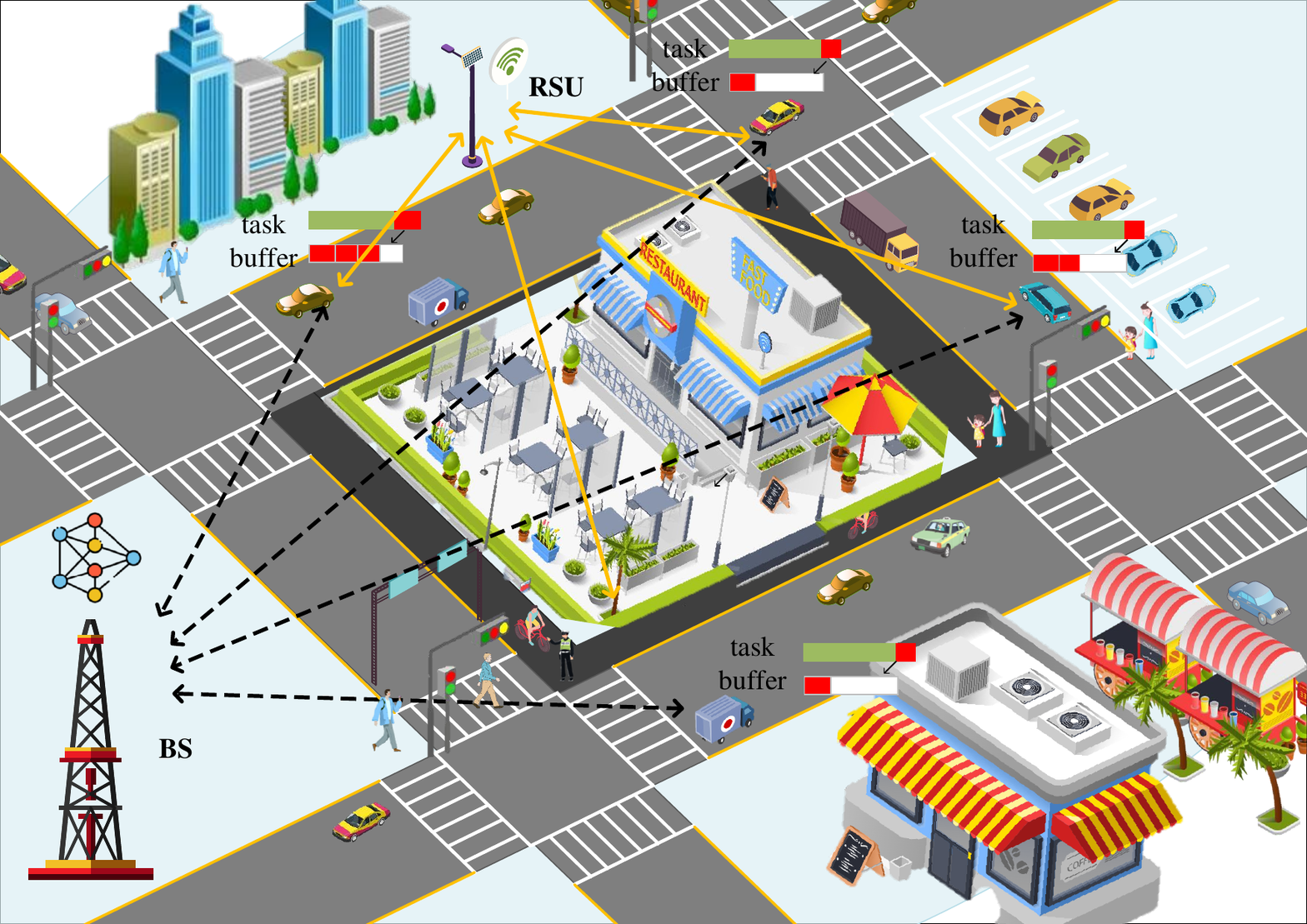}
	\caption{\textcolor{red}{System scenario}}
	\label{fig1}
\end{figure}

In this section, we will introduce the system scenario and models in this paper. As shown in Fig. \ref{fig1}, the scenario is an urban VEC traffic system containing four intersections. In this scenario, a BS is located at the corner, an RSU is deployed by the roadside, and $N$ vehicles are moving on the roads. We assume that all four intersections are within the coverage area of both BS and RSU. When vehicle $n$, $n \in [1, N]$, arrives at an intersection, it chooses to turn left, turn right, or go straight through the intersection with probabilities of $\psi_1$, $\psi_2$, and $\psi_3$, respectively. The neural network configured on the BS is used for decision support. The RSU  has fixed computing resource to parallel process tasks offloaded from vehicles \cite{7,8}.

The entire process consists of two stages: offloading-based federated SSL and DRL-based resource allocation. These two stages are interdependent yet mutually influential. The stage of offload-based federated SSL consists $E^{max}$ episodes, and each episode $e$, $e\in[1,E^{max}]$, is divided into $R^{max}$ rounds. Similarly, the stage of DRL-based resource allocation also consists $E^{max}$ episodes, and each episode $e$ is divided into $S^{max}$ slots \cite{9,10,11}. Each round of FL corresponds to a slot, and the duration is $T$. However, when slot $t$ reaches to $S^{max}$, the episode $e$ enter to the next episode $e+1$, and slot $t$ resets to zero. Meanwhile, the round $r$ continues to accumulate until the entire algorithm is completed. Thus, the relationship between round $r$ and slot $t$ in episode $e$ is described as 
\begin{equation}
	\textcolor{red}{r = \left(e - 1\right) \cdot S^{max} + t, \quad e \in [1, E^{max}] \text{ and } t \in [1, S^{max}]}
	\label{eq51}
\end{equation}

Following, we will introduce the computation model, channel model and transmission model in each round/slot.

\subsection{Computation model}

During the training process, the energy consumed for performing one training iteration on vehicle $n$ is calculated as follows
\begin{equation}
	\textcolor{red}{E_{t,L}^{n,comp} = p_{t,L}^{n,comp} T_{t,L}^{n,comp}}
	\label{eq1}
\end{equation}where $p_{t,L}^{n,comp}$, $T_{t,L}^{n,comp}$, and $E_{t,L}^{n,comp}$ represent the computing power, delay, and energy consumption, respectively, for vehicle $n$ to perform one local training iteration. The local CPU computation power $p_{t,L}^{n,comp}$ of vehicle $n$ in slot $t$ can be calculated using the Dynamic Voltage and Frequency Scaling (DVFS) method as follows
\begin{equation}
	\textcolor{red}{p_{t,L}^{n,comp}=\kappa\left(f_{t,L}^{n,comp}\right)^3}
	\label{eq2}
\end{equation}where $\kappa$ is the effective switching capacitance, which depends on the chip architecture. $f_{t,L}^{n,comp}\in\left[f^{min},\ f^{max}\right]$, where $f^{min}$ and $f^{max}$ represent the minimum and maximum CPU frequency, respectively \cite{hxiao2021}.

To calculate the computation delay of vehicle $n$, let $c$ be the number of CPU cycles required to process a unit size of data. $Z$ is the size of the local data, and it is assumed that each vehicle has the same data size. Then, $cZ$ represents the number of CPU cycles needed for one iteration. Therefore, the computation delay $T_{t,L}^{n,comp}$ is calculated as follows
\begin{equation}
	\textcolor{red}{T_{t,L}^{n,comp}=\frac{cZ}{f_{t,L}^{n,comp}}}
	\label{eq3}
\end{equation}

By substituting Eqs. \eqref{eq2} and \eqref{eq3} into Eq. \eqref{eq1}, we obtain
\begin{equation}
	\textcolor{red}{E_{t,L}^{n,comp\ }=\kappa\left.\left(f_{t,L}^{n,comp\ }\right.\right)^2cZ}
	\label{eq4}
\end{equation}

Similarly, we can calculate the delay $T_{t,R}^{comp}$ and energy consumption $E_{t,R}^{comp}$ for each computation iteration at RSU side as
\begin{equation}
	\textcolor{red}{T_{t,R}^{comp}=\frac{cZ}{f_{t,R}^{comp}}}
	\label{eq41}
\end{equation}
\begin{equation}
	\textcolor{red}{E_{t,R}^{comp\ }=\kappa\left.\left(f_{t,R}^{comp\ }\right.\right)^2cZ}
	\label{eq40}
\end{equation}where $f_{t,R}^{comp\ }$ is the CPU frequency for computing of RSU in slot $t$.

\subsection{Channel model}
The vehicles involved in training use Orthogonal Frequency Division Multiple (OFDM) technology to communicate with the RSU \cite{qwu2023,12}, so vehicle-to-vehicle interference is not considered during communication. Each vehicle uses the same channel model in the same slot, and the channel model is updated as the slot $t$ advances to slot $t+1$ \cite{13,14}. Therefore, in slot $t$, the channel gain can be calculated as
\begin{equation}
	\textcolor{red}{h_{t,R}^{n,trans}\left[\text{dis}(R,n)\right]=\alpha_t^n\left[\text{dis}(R,n)\right]m_t^n}
	\label{eq5}
\end{equation}where $m_t^n$ denotes small-scale fading, which follows an exponential distribution with a unit mean, i.e., $m_t^n \sim \text{Exp}(1)$. $\text{dis}(R,n)$ represents the distance between vehicle $n$ and the RSU, and measured in meters. $\alpha_t^n$ represents the effect of large-scale fading on the signal, including path loss $\mathcal{P}$ and shadow fading $\mathcal{S}$. Thus, $\alpha_t^n$ can be calculated as
\begin{equation}
	\textcolor{red}{\alpha_t^n={10}^\frac{\mathcal{P}}{10}{10}^\frac{\mathcal{S}}{10}}
	\label{eq6}
\end{equation}where shadow fading $\mathcal{S}$ follows a normal distribution with a mean of $\mu$ and a standard deviation of $\sigma$, represented as $\mathcal{S} \sim \mathcal{N}\left(\mu,\sigma^2\right)$. In high-density urban environments, the path loss $\mathcal{P}$ can be calculated as
\begin{equation}
	\textcolor{red}{\mathcal{P} =128.1+37.6\log\left[\frac{\text{dis}(R,n)}{1000}\right]}
	\label{eq73}
\end{equation}

Therefore, the information transmission rate $	R_{t,R}^n$ between the vehicle $n$ and the RSU can be calculated as \cite{sluo2020}
\begin{equation}
	\textcolor{red}{R_{t,R}^n = B^n \log_2 \left[ 1 + \frac{p_{t,R}^n h_{t,R}^{n,trans} \text{dis}(R,n)}{N_0} \right]}
	\label{eq8}
\end{equation}where $N_0$ represents the noise power, and $B^n$	denotes the bandwidth occupied by vehicle $n$ when transmitting data to RSU.

\subsection{Transmission model}

If vehicle $n$ offloads partial iterations task to the RSU, during the transmission process, the energy consumption $E_{t}^{n,trans}$ can be calculated as
\begin{equation}
	\textcolor{red}{E_{t}^{n,trans}=p_{t,R}^n T_{t,R}^{n,trans}}
	\label{eq9}
\end{equation}where $ T_{t,R}^{n,trans}$ is the time required for vehicle $n$ to transmit data to RSU. Assuming each vehicle transmits a local model of the same size $D$, and the data transmitted by vehicle $n$ to RSU includes the untrained local model and local data. Therefore, the transmission delay $T_{t,R}^{n,trans}$ can be calculated as
\begin{equation}
	\textcolor{red}{T_{t,R}^{n,trans}=\frac{Z+D}{R_{t,R}^n}}
	\label{eq10}
\end{equation}
Thus, the transmission energy consumption $E_{t}^{n,trans}$ from vehicle $n$ to RSU can be calculated as 
\begin{equation}
	\textcolor{red}{E_{t}^{n,trans}=\frac{p_{t,R}^n(Z+D)}{R_{t,R}^n}}
	\label{eq100}
\end{equation}

\section{\textcolor{red}{Optimization problem}}
	\label{sec4}

When vehicles are performing local iterations task, if they are unable to complete all iterations locally due to local computing resource limitations, partial iterations task needs to be offloaded to RSU \cite{fan3}. This offloading process requires BS to use a DRL algorithm to allocate assignment ratios, which determines the computing resource of the RSU allowed for each vehicle. The total iterations are then allocated between local training and training on the RSU. This optimization aims to enhance resource utilization and task completion efficiency. Next, we will delve into the specific methods of task allocation.

\subsection{Task allocation}
\label{Task allocation}

Within slot $t$, vehicle $n$ is able to determine the total local training iterations $N_t^n$, i.e.,
\begin{equation}
	\textcolor{red}{N_t^n=\frac{T-t^{max}}{T_{t,L}^{n,comp}}}
	\label{eq11}
\end{equation}where $t^{max}$ represents the maximum time used for data transmission. Thus, $T-t^{max}$ indicates the total time available for computing within a slot. In practical, the CPU typically handles multiple tasks simultaneously to improve CPU utilization. These tasks consume some CPU computing resources, resulting in the actual time available for local iterations being less than $T$. We denote the true training time as $T'$. Therefore, the actual total local training iterations ${N_t^n}^{\ast}$ can be calculated as
\begin{equation}
	\textcolor{red}{{N_t^n}^{\ast}=\frac{T'-t^{max}}{T_{t,L}^{n,comp}}}
	\label{eq12}
\end{equation}

Substitute Eq. \eqref{eq3} into Eq. \eqref{eq11} and Eq. \eqref{eq12}, the final results can be obtained, respectively.
\begin{equation}
	\textcolor{red}{N_t^n=\frac{(T-t^{max})f_{t,L}^{n,comp}}{cZ}}
	\label{eq111}
\end{equation}
\begin{equation}
	\textcolor{red}{{N_t^n}^{\ast}=\frac{(T'-t^{max})f_{t,L}^{n,comp}}{cZ}}
	\label{eq121}
\end{equation}

In this paper, we study a binary partial offloading problem, where a binary variable $g_t^n$ is used to indicate whether vehicle $n$ needs to offload partial task to the RSU. $g_t^n$ is related to the assignment ratio $q_t^n$ and the assignment ratio threshold $q_0$. When $q_t^n$ is less than the assignment ratio threshold $q_0$, it indicates that vehicle $n$ will not offload partial task to RSU and reset $q_t^n=0$; otherwise, it will offload it. Therefore, $g_t^n$ can be expressed as
\begin{equation}
	\textcolor{red}{g_t^n = 
	\begin{cases} 
		1, & \text{if } q_t^n \geq q_0 \\
		0, & \text{if } q_t^n < q_0
	\end{cases}}
	\label{eq13}
\end{equation}
\begin{figure}
	\centering
	\includegraphics[scale=0.4, trim=0.1cm 2.5cm 0.1cm 2.5cm, clip]{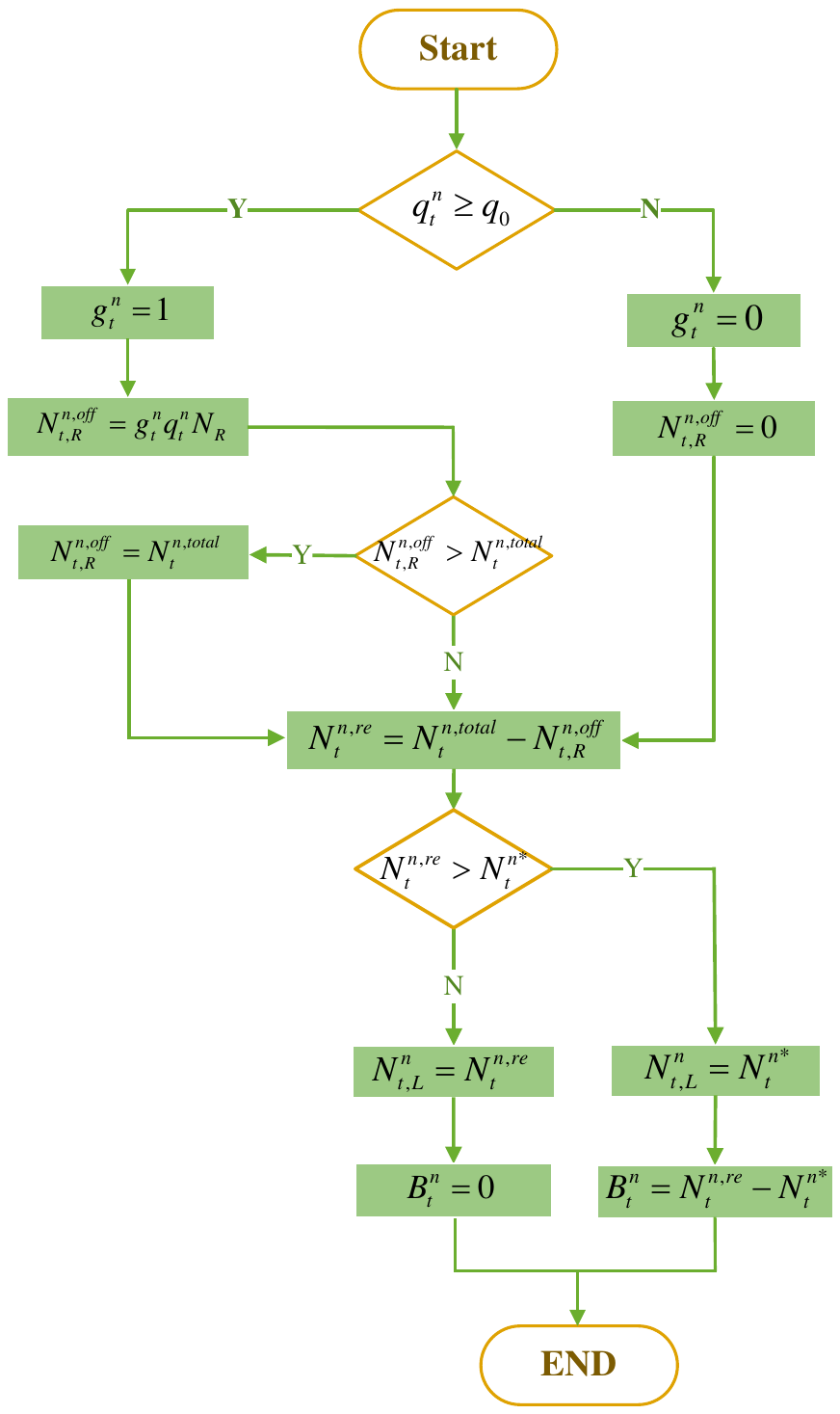}
	\caption{\textcolor{red}{Task allocation flowchart}}
	\label{fig2}
\end{figure}
We denote the total task that vehicle $n$ needs to handle in slot $t$ as $N_t^{n,total}$, which includes the sum of the total training iterations $N_t^n$ in slot $t$ and the remaining unprocessed training iterations $B_{t-1}^n$ from the previous slot. When vehicle $n$ offloads partial task to RSU, the total task $N_t^{n,total}$ will be divided into local training iterations $N_{t,L}^{n}$ and offloaded training iterations $N_{t,R}^{n,off}$ according to the allocated assignment ratio $q_t^n$. Specifically, RSU first determines the total training iterations $N_R$ in RSU side as follows
\begin{equation}
	\textcolor{red}{N_R=\frac{T-t^{max}-T_{t,R}^{n,trans}}{T_{t,R}^{comp}}}
	\label{eq14}
\end{equation}

Thus, the expected offloaded training iterations $N_{t,R}^{n,off}$ for RSU can be calculated as
\begin{equation}
	\textcolor{red}{N_{t,R}^{n,off}=g_t^n  q_t^nN_R}
	\label{eq15}
\end{equation}

If $N_{t,R}^{n,off}$ exceeds $N_t^{n,total}$, it indicates that the expected offloading task is too large, surpassing the total task that need to be processed locally. In this case, the actual offloaded training iterations are recorded as $N_t^{n,total}$. Thus, $N_{t,R}^{n,off}$ can be updated as
\begin{equation}
	\textcolor{red}{N_{t,R}^{n,off} \gets \min \left(N_{t,R}^{n,off}, N_t^{n,total} \right)}
	\label{eq17}
\end{equation}

So far, we have obtained the actual offloading iterations $N_{t,R}^{n,off}$. Following, we will calculate the number of local iterations $N_{t,L}^n$.

The remaining portion of $N_t^{n,total}$ denoted as $N_t^{n,re}$, which represents the number of iterations that need to be processed locally on the vehicle $n$, and can be calculated as
\begin{equation}
	\textcolor{red}{N_t^{n,re}=N_t^{n,total}-N_{t,R}^{n,off}}
	\label{eq18}
\end{equation}

For the local computing capability of the vehicle $n$, its actual total number of training iterations is ${N_t^n}^{\ast}$. When $N_t^{n,re}$ does not exceed ${N_t^n}^{\ast}$, vehicle $n$ can handle all remaining task locally, i.e., $N_{t,L}^{n}=N_t^{n,re}$. When $N_t^{n,re}$ exceeds ${N_t^n}^{\ast}$, it means vehicle $n$ cannot handle all remaining task locally, and the maximum iterations can be deal with locally are ${N_t^n}^{\ast}$. Meanwhile, the unprocessed portion is placed in the local buffer $B_t^n$. We abbreviate $N_{t,L}^{n}$ as
\begin{equation}
	\textcolor{red}{N_{t,L}^{n}=\min \left(N_{t}^{n,re}, {N_t^n}^{\ast} \right)}
	\label{eq19}
\end{equation}and $B_t^n$ is expressed as
\begin{equation}
\textcolor{red}{	B_t^n = 
	\begin{cases} 
		0, & \text{if }   N_t^{n,re}\leq {N_t^n}^{\ast} \\
		N_t^{n,re}-{N_t^n}^{\ast}, & \text{if } N_t^{n,re} > {N_t^n}^{\ast}
	\end{cases}}
	\label{eq20}
\end{equation}

The specific process of task allocation is as shown in Fig. \ref{fig2}, and the involved symbols for vehicle $n$ in slot $t$ are listed in Table \ref{tab1}.
\begin{table}[h]\footnotesize
	\centering
	\begin{threeparttable}
		\setlength{\abovecaptionskip}{0pt}
		\caption{Symbols and meanings.}
			\label{tab1}
		\begin{tabular}{p{0.8cm} p{7.5cm}}
			\hline
			\textbf{Symbol} & \textbf{Meaning}\\
			\hline
			$p_{t,B}^n$ & Transmission power from vehicle to BS \\ 
			$p_{t,R}^n$ & Transmission power from vehicle to RSU \\
			$p_{t,L}^{n,comp}$ & Computing power for vehicle \\
			$f_{t,L}^{n,comp}$ & CPU frequency for vehicle \\
			$f_{t,R}^{comp}$ & CPU frequency for RSU \\
			$N_t^n$ & Expected total number of training iterations \\
			${N_t^n}^\ast$ & Actual total number of training iterations \\
			$N_{t,L}^n$ & Number of local iterations \\
			$N_{t,R}^{n,off}$ & Offloaded training iterations \\
			$N_R$ & Total training iterations in RSU \\
			$N_t^{n,total}$ & Total task \\
			$N_t^{n,re}$ & Remaining portion of total task\\
			$T_{t,L}^{n,comp}$ & Delay for each computation iteration at vehicle \\
			$T_{t,R}^{comp}$ & Delay for each computation iteration at RSU \\
			$T_{t,R}^{n,trans}$ & Transmission delay from vehicle to RSU \\
			$h_{t,R}^{n,trans}$ & Channel gain \\
			$q_t^n$ & Assignment ratio \\
			$m_t^n$ & Small-scale fading \\	
			$\alpha_t^n$ & Large-scale fading \\
			$E_{t,L}^{n,comp}$ & Energy consumption for each computation iteration at vehicle \\
			$E_{t,R}^{comp}$ & Energy consumption for each computation iteration at RSU \\
			$E_t^{n,trans}$ & Energy consumption for transmission from vehicle to RSU \\
			$R_{t,R}^n$ & Information transmission rate between the vehicle and RSU \\
			$g_t^n$ & If offloading \\
			$B_{t-1}^n$ & Remaining unprocessed training iterations from the previous slot \\
			\hline

\end{tabular}
\end{threeparttable}
\end{table}

\subsection{Energy consumption}
Increasing the number of local iterations typically enhances classification accuracy, but it also leads to increased energy consumption per round of FL. To find a balance between classification accuracy and energy consumption, we shall formulate an optimization model to find its solution.

Due to the self-interested nature of individual vehicles, their decisions often aim to maximize their own benefits. However, this self-interested behavior can lead to imbalances in system resource utilization efficiency and energy consumption. In contrast, a key advantage of centralized allocation by BS is its ability to balance overall system resource utilization efficiency and energy consumption. In this section, we propose an optimization algorithm aimed at minimizing the energy consumption of the entire system. The energy consumption is divided into three parts: RSU computing energy consumption $E_{t,R}^{n}$, transmission energy consumption $E_t^{n,trans}$ from vehicle $n$ to RSU, and local computing energy consumption $E_{t,L}^{n}$. $E_t^{n,trans}$ can be calculated according to Eq. \eqref{eq100}. $E_{t,R}^{n}$ and $E_{t,L}^{n}$ are represented as, respectively
\begin{equation}
	\textcolor{red}{E_{t,R}^n=E_{t,R}^{comp} N_{t,R}^{n,off}}
	\label{eq21}
\end{equation}
\begin{equation}
	\textcolor{red}{E_{t,L}^n=E_{t,L}^{n,comp} N_{t,L}^{n}}
	\label{eq22}
\end{equation}

Therefore, the total energy consumption $E_t^{n,total}$ for each vehicle $n$ in slot $t$ can be expressed as			
\begin{equation}
	\textcolor{red}{E_t^{n,total}=g_t^n E_{t,R}^n+E_{t,L}^n+g_t^n E_t^{n,trans}}
	\label{eq23}
\end{equation}			

\subsection{Problem formulation}
To achieve efficient task offloading and resource allocation, it is essential for BS to consider the long-term impact of their actions during the decision-making process. Therefore, we aim to minimize the sum of the energy consumption of all vehicles in each slot. Let $\mathbb{P}=\left\{p_{t,R}^1,p_{t,R}^2,\cdots p_{t,R}^n,\cdots p_{t,R}^N\right\}$, $\mathbb{F}=\left\{f_{t,L}^{1,comp},f_{t,L}^{2,comp},\cdots f_{t,L}^{n,comp},\cdots f_{t,L}^{N,comp}\right\}$ and $\mathbb{Q}=\left\{q_t^1,q_t^2,\cdots q_t^n,\cdots q_t^N\right\}$. Then, we can formulate the optimization problem in slot $t$ as follows	
\begin{equation}
	\textcolor{red}{\underset{\mathbb{P}, \mathbb{F}, \mathbb{Q}}{\min} : \sum_{n=1}^{N} E_{t}^{n,total}}
	\label{eq24}
\end{equation}
\hspace{5em}\text{s.t.}
\begin{subequations}
	\begin{equation}
		\textcolor{red}{g_t^n \in \left\{0,1 \right\}, \forall n \in [1,N]\tag{\ref{eq24}{a}} \label{eq24a}}
	\end{equation}
	\begin{equation}
		\textcolor{red}{p^{min} \leq p_{t,R}^n \leq p^{max}, \forall n \in [1,N] \tag{\ref{eq24}{b}} \label{eq24b}}
	\end{equation}
	\begin{equation}
		\textcolor{red}{f^{min} \leq f_{t,L}^{n,comp} \leq f^{max}, \forall n \in [1,N]\tag{\ref{eq24}{c}} \label{eq24c}}
	\end{equation}
	\begin{equation}
		\textcolor{red}{0 \leq q_t^n \leq 1, \forall n \in [1,N]\tag{\ref{eq24}{d}} \label{eq24d}}
	\end{equation}
	\begin{equation}
		\textcolor{red}{\sum_{n=1}^{N} q_t^n \leq 1\tag{\ref{eq24}{e}} \label{eq24e}}
	\end{equation}
\end{subequations}

Eq. \eqref{eq24a} signifies that the offloading task is  formulated as a binary decision problem, i.e., whether to offload or not. $p^{min}$ and $p^{max}$ is the the minimum and maximum transmission power.  Eq. \eqref{eq24b} restricts the transmission power to the range between $p^{min}$ and $p^{max}$, while  Eq. \eqref{eq24c} limits the CPU frequency of vehicle $n$ to the range between $f^{min}$ and $f^{max}$. Eqs. \eqref{eq24d} and \eqref{eq24e} constrain the assignment ratio, ensuring that the number of offloaded tasks does not exceed the computing capacity of RSU.

\subsection{Evaluation metrics}

Evaluation metrics quantify the system's performance, ensuring effective and objective assessment and optimization of task offloading and resource allocation scheme. These metrics enable fair comparison across different algorithms, thereby enhancing the overall system performance and efficiency. In this section, we will introduce the evaluation metrics used in our paper.

\subsubsection{Overload ratio}

After the BS allocates the assignment ratio $q_t^n$ for vehicle $n$ in slot $t$, the iterations number of offloading task $N_t^{n,off}$ assigned to vehicle $n$ might exceed the total number of tasks $N_t^{n,total}$ that the vehicle $n$ needs to process in slot $t$ \cite{jyu2022}. We denote the excess part as the overload $O_t^n$, which can be calculated as
\begin{equation}
	\textcolor{red}{O_t^n = \max \left(0, N_{t,R}^{n,off} - N_t^{n,total} \right)}
	\label{eq16}
\end{equation}

We use the overload ratio $\mathcal{R}_0$ to describe the average severity of the overload behavior across all vehicles over $S^{max}$ slots, that is,
\begin{equation}
	\textcolor{red}{\mathcal{R}_0=\frac{1}{NS^{max}} \left( \sum_{n=1}^N\sum_{t=1}^{S^{max}}{\frac{O_t^n}{N_t^{n,total}}} \right)}
	\label{eq25}
\end{equation}

\subsubsection{Offloading efficiency}
As described above, when the assignment ratio $q_t^n$ is below a specific threshold $q_0$, vehicle $n$ will not perform offloading operations. Therefore, we define offloading efficiency $\eta_{q_0}$ as a metric to measure the system's capability to offload tasks, i.e.,
\begin{equation}
	\textcolor{red}{\eta_{q_0}=\frac{1}{NS^{max}}\sum_{n=1}^{N}\sum_{t=1}^{S^{max}}\left(\frac{N_t^{n,off}}{N_t^{n,total}}\times100\%\right)}
	\label{eq26}
\end{equation}

\section{DRL-based solution}
\label{sec5}

The problem of task offloading and resource allocation can be addressed with a DRL algorithm. To this end, we will utilize DRL to find the optimal task offloading and resource allocation scheme. The SAC algorithm is chosen due to its robustness and efficiency in handling continuous action spaces and its ability to balance exploration and exploitation effectively. SAC is an off-policy DRL algorithm designed to enhance the stability of learning and the exploration of the policy. It encourages exploration through the maximum entropy framework, preventing premature convergence to suboptimal solutions. Additionally, SAC uses DQN to reduce overestimation bias in value function approximation, further improving the stability. Compared to other DRL algorithms, SAC introduces an entropy term into the objective function to encourage BS to continuously explore new policies. The presence of this entropy term effectively prevents BS from prematurely converging to a local optimal solution during policy iteration, thereby preventing training failure.

We employ SAC algorithm with the objective of minimizing the overall system's energy consumption, aiming to find an optimal resource allocation scheme. Next, we will introduce this algorithm.

For the decision-making process in each slot $t$, it includes state, action, reward, and solution. We define the state, action, reward, and solution in slot $t$ as follows

\subsection{State}

In this case, BS is able to analysis state and experiences during the process of decision-making. The system state consists of two elements: the set of gain information $\mathbb{G}=\left\{h_{t,R}^{1,trans},h_{t,R}^{2,trans},\cdots h_{t,R}^{n,trans},\cdots h_{t,R}^{N,trans}\right\}$, the set of vehicle velocities $\mathbb{V}=\left\{v_t^1,v_t^2,\cdots v_t^n,\cdots v_t^N\right\}$ and the total task $\mathbb{N}=\left\{N_{t}^{1,total},N_{t}^{2,total},\cdots N_{t}^{n,total},\cdots N_{t}^{N,total}\right\}$. Here, the gain information $ h_t^n$ is determined based on the current position of vehicle $n$, and the velocity $v_t^n$, $v_t^n \in [v^{min}, v^{max}]$, remains constant within each slot and changes at the beginning of next slot. $v^{min}$ and $v^{max}$ respectively denote the minimum and maximum values of vehicle velocity. Therefore, the state can be expressed as
\begin{equation}
	\textcolor{red}{s_t=\left( \mathbb{G}, \mathbb{V}, \mathbb{N}\right)}
	\label{eq28}
\end{equation}

\subsection{Action}

After obtaining the state, the BS makes decisions based on SAC algorithm for slot $t$. We set the action as three discrete variables: the set of powers $\mathbb{P}=\left\{p_{t,R}^1,p_{t,R}^2,\cdots p_{t,R}^n,\cdots p_{t,R}^N\right\}$ when the vehicles send data to RSU, the set of CPU computation frequencies $\mathbb{F}=\left\{f_{t,L}^{1,comp},f_{t,L}^{2,comp},\cdots f_{t,L}^{n,comp},\cdots f_{t,L}^{N,comp}\right\}$, and the set of assignment ratios $\mathbb{Q}=\left\{q_t^1,q_t^2,\cdots q_t^n,\cdots q_t^N\right\}$. Thus, the action in slot $t$ can be expressed as
\begin{equation}
\textcolor{red}{	a_t=\left( \mathbb{P}, \mathbb{F}, \mathbb{Q}\right)}
	\label{eq29}
\end{equation}

\subsection{Reward}
Our objective is to minimize energy consumption, overload amount, and the remaining local iterations in the buffer within each slot. Thus, in slot $t$, the reward consists of three negative components: the total energy consumption $E_t^{n,total}$, the overload amount $O_t^n$, and the remaining local iterations $B_t^n$ at the end of slot $t$. The reward function can be calculated as
\begin{equation}
\textcolor{red}{	r_t=- \sum_{n=1}^{N}{\left(\vartheta_1E_t^{n,total}+\vartheta_2O_t^n+\vartheta_3B_t^n\right)}}
	\label{eq30}
\end{equation}where $\vartheta_1$, $\vartheta_2$, and $\vartheta_3$ represent weight of the penalty term.

\subsection{SAC-based task offloading and resource allocation solution in federated SSL}
The SAC algorithm bases on the maximum entropy principle, and it involves five networks: one actor network $\varsigma$, two critic networks $\xi_1$ and $\xi_2$, and two target networks $\varpi_1$ and $\varpi_2$. At the start of the scheme, the BS holds the parameters of the global model, while each vehicle retains a local model that has the same structure as the global model.
Following, we will introduce the complete SAC-based federated SSL solution for task offloading and resource allocation.

\textbf{\emph{Step 1, initialization}}: When episode $e=1$, if in the first round/slot, the BS randomly initializes the global model, as well as the actor and critic networks within the SAC algorithm, and assigns the parameters of the critic networks to two target networks.


\textbf{\emph{Step 2, task offloading and training}}: In episode $e$, before slot $t$ starts, determine the relationship between round $r$ and slot $t$ according to Eq. \eqref{eq51} firstly. BS stores the parameters of a global model $\theta_{{r-1}}^{\text{global}}$ aggregated in previous round $r-1$, along with the five networks of SAC algorithm. If in episode $e$, slot $1$, the parameter of five networks come from episode $e-1$, slot $S^{max}$. Vehicles also store $\mathcal{Z}$ image data, denoted as local data, and a local model which has the same structure as global model. Then, BS employs SAC algorithm to allocate $p_{t,R}^n$, $f_{t,L}^{n,comp}$, and $q_t^n$ for vehicle $n$. Next, vehicle $n$ downloads the global model $\theta_{{r-1}}^{\text{global}}$ from BS and assigns the parameters of global model to local model $\theta_{{r}}^{\text{local},n}$, along with the transmission power $p_{t,R}^n$, $f_{t,L}^{n,comp}$, and $q_t^n$ allocated by BS for vehicle $n$. Based on the downloaded CPU frequency $f_{t,L}^{n,comp}$, the total training time $T$, and true training time $T'$, $N_t^n$ and ${N_t^n}^{\ast}$ can be calculated by Eq. \eqref{eq111} and Eq. \eqref{eq121}. Then, conducting task allocating to obtain $N_{t,L}^{n}$ and $N_{t,R}^{n,off}$ according to Section \ref{Task allocation}. If the downloaded assignment ratio $q_t^n$ is greater than the offloading threshold $q_0$, the vehicle $n$ immediately sends an offloading request to RSU with the downloaded transmission power $p_{t,R}^n$. This request includes local data $\left\{x_{r}^{1,n},\ x_{r}^{2,n},\ \ldots x_{r}^{i,n},\ldots x_{r}^{\mathcal{Z},n}\right\}$, local model $\theta_{{r}}^{\text{local},n}$, and assignment ratio $q_t^n$.


Meanwhile, vehicle $n$ also performs local training with local data and local model for $N_{t,L}^{n}$ iterations. The training process is a federated SSL. For each iteration, first, input each local data $x_{r}^{i,n}$, $x_{r}^{i,n}\in\left\{x_{r}^{1,n},\ x_{r}^{2,n},\ \ldots x_{r}^{i,n},\ldots x_{r}^{\mathcal{Z},n}\right\}$, into two different data augmentation methods, $\pi_1(\cdot)$ and $\pi_2(\cdot)$. The function $\pi_1(\cdot)$ applies a horizontal flip to the image with a 50\% likelihood, and converts the image to grayscale with a 20\% likelihood. Meanwhile, $\pi_2(\cdot)$ adjusts the brightness of image, saturation, and hue with an 80\% likelihood, and then converts the image to grayscale with a 40\% likelihood. These two different augmentations are adopted in the FLSimCo, so the features of image can be efficiently utilized.

The remaining images $x_{r}^{j,n},j\in\left[1,\mathcal{Z}\right]\ \text{and}\ j\neq i$ are denoted as negative samples. Feeding augmented images $x_{r}^{i,n}$ and $x_{r}^{j,n}$ into a function $\chi$, which is composed of an encoder, two dense neutral networks and a ReLU function. The output of the $\chi$ includes anchor sample $q_{r}^{i,{n}}$, positive sample $k_{{r}}^{i,{n}}$, and encoded negative samples $k_{{r}}^{j,{n}}$, respectively, i.e., 

\begin{equation}
	\textcolor{red}{q_{r}^{i,n}=\chi\left\{\pi_1\left(x_{r}^{i,n}\right)\right\},\ i\in\left[1,\mathcal{Z}\right]}
	\label{eq50}
\end{equation}
\begin{equation}
	\textcolor{red}{k_{r}^{i,n}=\chi\left\{\pi_2\left(x_{r}^{i,n}\right)\right\},\ i\in\left[1,\mathcal{Z}\right]}
	\label{eq60}
\end{equation}
\begin{equation}
	\textcolor{red}{k_{r}^{j,n}=\chi\left\{x_{r}^{j,n}\right\},\ j\in\left[1,\mathcal{Z}\right],\ \text{and}\ j\neq i}
	\label{eq70}
\end{equation}

The loss can be calculated by Information Noise Contrastive Estimation (InforNCE) as \cite{aoord2019}
\begin{equation}
	\textcolor{red}{\begin{split}
		\mathcal{L}_{{q_{{r}}^{i,n}}|\tau={\tau_1}} =\frac{\exp{\left(\frac{q_{{r}}^{i,n}\cdot k_{{r}}^{i,n}}{\tau}\right)}}{\exp{\left(\frac{q_{{r}}^{i,n}\cdot k_{{r}}^{i,n}}{\tau}\right)}+\sum_{j=1}^{\mathcal{K}}\exp{\left(\frac{q_{{r}}^{i,n}\cdot k_{{r}}^{j,n}}{\tau}\right)}}
	\end{split}}
	\label{eq71}
\end{equation}where $\mathcal{K}$ is the number of negative samples, and $\tau$ is temperature hyper-parameter \cite{khou2020}. Thus, the dual temperature loss used by FLSimCo can be calculated as

\begin{equation}
\textcolor{red}{	\begin{split}
		\mathcal{L}_{q_{{r}}^{i,n}}^{\text{DT}}=\text{sg}\left[\frac{1-\mathcal{L}_{{q_{{r}}^{i,n}}|\tau={\tau_2}}}{1-\mathcal{L}_{{q_{{r}}^{i,n}}|\tau={\tau_1}}}\right] \times \log\mathcal{L}_{{q_{{r}}^{i,n}}|\tau={\tau_1}}
	\end{split}}
	\label{eq7}
\end{equation}
where $\text{sg}[\cdot]$ indicates the stop gradient. $\tau_1$ and $\tau_2$ are different values of temperature hyper-parameter $\tau$. We use the Stochastic Gradient Descent (SGD) method to update the local model \cite{czhang2022dualtemp}, i.e.,
\begin{figure*}[htbp]
	\centering
	\includegraphics[scale=0.63, trim=0.2cm 2cm 0.2cm 2cm, clip]{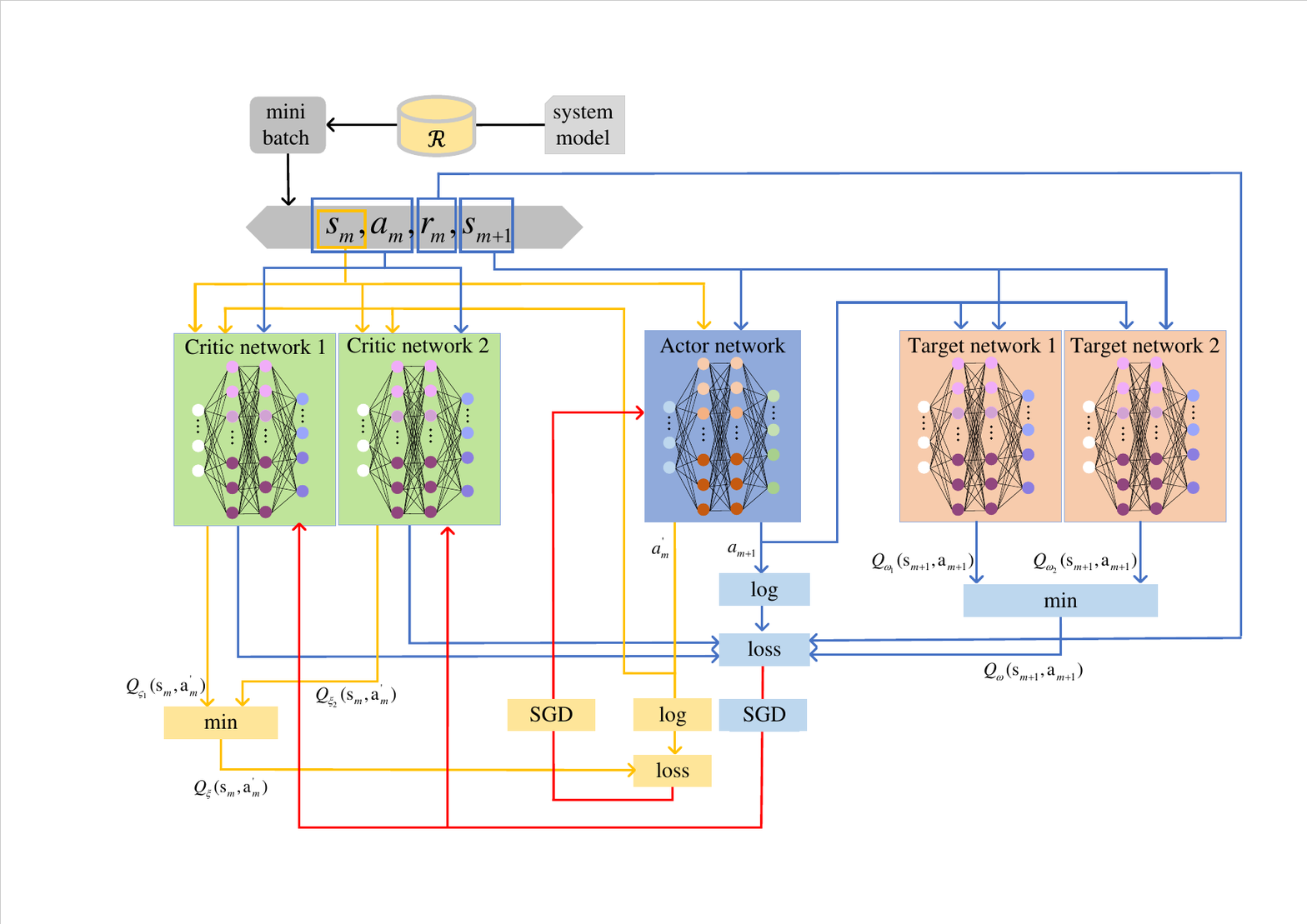}
	\caption{\textcolor{red}{The process of updating actor and critic networks of SAC algorithm}}
	\label{fig8}
\end{figure*}

\begin{equation}	
	\textcolor{red}{\theta_{{r,L}}^{\text{local},n} = \theta_{{r}}^{\text{local},n}-\eta^{r}\left[ \nabla_{q_{{r}}^{i,n}}\left(\frac{1}{\mathcal{Z}}\sum_{i=1}^{\mathcal{Z}} \mathcal{L}_{q_{{r}}^{i,n}}^{\text{DT}}\right)\right]}
\end{equation}where $\eta^{r}$ is the learning rate and $\theta_{{r,L}}^{\text{local},n}$ is the model trained locally by vehicle $n$ in round $r$. Noting that, during the process of local training, vehicle $n$ also captures images $\left\{x_{r+1}^{1,n},\ x_{r+1}^{2,n},\ \ldots x_{r+1}^{i,n},\ldots x_{r+1}^{\mathcal{Z},n}\right\}$, as local data, for next round.

At the same time, upon receiving the vehicle's offloading request, RSU performs training for $N_{t,R}^{n,off}$ iterations using the local data and local model upload by vehicle $n$ to generate a new model $\theta_{{r,R}}^{\text{local},n}$, similar with the training process as vehicle locally training.
After the training is completed, the RSU sends the trained model $\left\{\theta_{{r,R}}^{\text{local},1},\ \theta_{{r,R}}^{\text{local},2},\ \ldots \theta_{{r,R}}^{\text{local},n},\ldots \theta_{{r,R}}^{\text{local},N}\right\}$ back to each vehicle. Based on $N_{t,L}^{n}$ and $N_{t,R}^{n,off}$, the total energy consumption $E_t^{n,total}$ can be calculated according to Eq. \eqref{eq23}.

\textbf{\emph{Step 3, uploading}}: After receiving all trained model sent by RSU, vehicles upload trained models $\left\{ \theta_{r,L}^{local,1},\theta_{r,L}^{local,2},\cdots \theta_{r,L}^{local,n},\cdots \theta_{r,L}^{local,N} \right\}$, $\left\{ \theta_{r,R}^{local,1},\theta_{r,R}^{local,2},\cdots \theta_{r,R}^{local,n},\cdots \theta_{r,R}^{local,N} \right\}$, velocities  $\left\{ v_{t+1}^1,v_{t+1}^2,\cdots v_{t+1}^n,\cdots v_{t+1}^N \right\}$ for slot $t+1$, total energy consumption $\left\{ E_t^{1,total},E_t^{2,total},\cdots E_t^{n,total},\cdots E_t^{N,total} \right\}$, and unprocessed training iterations $\left\{B_t^1,B_t^2,\cdots B_t^n,\cdots B_t^N \right\}$ to BS, respectively.

\textbf{\emph{Step 4, aggregating and updating}}: Upon receiving the models from all vehicles, the BS aggregates received models followed as
\begin{equation}
\textcolor{red}{	\theta_{r}^{\text{global}}=\frac{\sum_{n=1}^{N}\left(\theta_{r,L}^{\text{local},n}+\theta_{r,R}^{\text{local},n}\right)}{2N}}
	\label{eq133}
\end{equation}where $\theta_{r}^{\text{global}}$ is the parameters of new global model in round $r$. 
Simultaneously, the reward is calculated by Eq. \eqref{eq30}, and the network is updated accordingly. Next, we will introduce the process of updating the SAC networks.

The core idea of the SAC algorithm is to balance the exploration-exploitation trade-off by increasing the stochasticity of policy based on entropy. The objective function $J(\pi_{\varsigma})$ of SAC can be expressed as the weighted sum of expected reward and entropy, i.e.,
\begin{equation}
\textcolor{red}{	J(\pi_{\varsigma}) = E_{(s_{t}, a_{t}) \sim \pi_{\varsigma}}\\\left[ \gamma^{t-1}r_t \right. 
	\left. + \beta H(\pi_{\varsigma}\left(\cdot | s_{t}\right)) \right]}
	\label{eq31}
\end{equation}where $H \left( \pi_{\varsigma}\left(\cdot | s_{t}\right) \right)$ denotes the entropy of actor network policy $\pi_{\varsigma}$ in state $s_t$. $\gamma \in [0,1] $ is the discount factor, and $\beta$ is the temperature parameter that adjusts the randomness of the optimal policy, balancing the importance of the expected reward and the entropy.	

Our goal is to maximize $J(\pi_{\varsigma})$, and upon achieving maximum $J(\pi_{\varsigma})$, the corresponding policy $\pi_{\varsigma}$ is denoted as $\pi_{\varsigma}^{\ast}$. Using $\pi_{\varsigma}^{\ast}$, we can adjust $\beta$ to obtain the optimal solution under state $s_t$, which is represented as $\beta^{\ast}$. Thus, $\beta^{\ast}$ can be expressed as
\begin{equation}
\textcolor{red}{	\beta^\ast = \arg\max_\beta J(\pi_{\varsigma}^{\ast})}
	\label{eq32}
\end{equation}

BS obtains the initial state $s_t$ from the system, then selects the action $a_t$ based on the policy of actor network $\pi_{\varsigma}$. After executing action $a_t$, the state $s_t$ transitions to a new state $s_{t+1}$ and receives the corresponding reward $r_t$. These interaction data are stored in the replay buffer $R$ in the form of tuples $(s_t, a_t, r_t, s_{t+1})$. The replay buffer $R$ allows for random sampling of these tuples, breaking the correlation between samples during training.
When the number of tuples stored in the replay buffer $R$ reaches to batch size $\mathcal{B}$, begin to update five networks of SAC algorithm. 

Randomly sample $M$ tuples from the replay buffer $R$, referred to as a mini batch. For each tuple $(s_m, a_m, r_m, s_{m+1})$ in the mini batch, where $m \in [1,M]$, feed $s_m$ into the actor network $\varsigma$. Based on the policy $\pi_\varsigma$ of the actor network $\varsigma$, a new action $a_m'$ is obtained. The loss function of $\beta$ can be calculated as 
\begin{equation}
\textcolor{red}{	\begin{split}
		\nabla_{\beta}J_{\beta} = - \nabla_{\beta}\ E_{a_m' \sim \pi_\varsigma} \left[  \beta\log \pi_\varsigma(a_m' | s_m) + \beta  \widetilde{H} \right]
	\end{split}}
	\label{eq33}
\end{equation}where, $ \widetilde{H}$ represents the dimensions of action $a_{t}$.

Next, as shown by the yellow line in Fig. \ref{fig8}, we start to update the actor network $\varsigma$. Firstly, feeding $(s_m, a_m')$ into the two critic networks $\xi_1$ and $\xi_2$, respectively, yielding the corresponding action value functions $Q_{\xi_1}(s_m, a_m')$ and $Q_{\xi_2}(s_m, a_m')$. The smaller of these two values is defined as $Q_{\xi}(s_m, a_m')$. Thus, by the SGD algorithm, the actor network $\varsigma$ is updated as follows
\begin{equation}
\textcolor{red}{	\begin{split}
		\nabla_\varsigma \mathcal{J}_\varsigma &= \nabla_\varsigma \left[ -\frac{1}{M} \sum_{m=1}^{M} \left( \beta \log \pi_\varsigma \left( a_m^\prime \middle| s_m \right) \right)^2 \right] \\
		&+ \nabla_\varsigma \left[ -\frac{1}{M} \sum_{m=1}^{M} \left( f_\varsigma(o; s_m) \times \nabla_{a_m'} \left( Q_{\xi}(s_m, a_m^\prime) \right. \right. \right. \\
		&\left. \left. \left. - \beta \log \pi_\varsigma \left( a_m^\prime \middle| s_m \right) \right) \right)^2 \right]
	\end{split}}
	\label{eq46}
\end{equation}where $o$ is noise sampled from a multivariate normal distribution, and $f_\varsigma(o; s_m)$ is a function utilized to re-parameterize the action $a_m^\prime$ \cite{thaarnoja2018}.

Following, as the blue line shown in Fig. \ref{fig8}, to update the critic networks, first input $(s_m, a_m)$ into both critic networks $\xi_1$ and $\xi_2$, and obtain the action value pairs $Q_{\xi_1}(s_m, a_m)$ and $Q_{\xi_2}(s_m, a_m)$, respectively. Meanwhile, input $s_{m+1}$ into the actor network $\varsigma$, and obtain $a_{m+1}$ under the policy $\pi_{\varsigma}$. To maintain randomness in the policy $\pi_{\varsigma}$ during training and explore a wider range of states and actions, the SAC algorithm introduces an entropy regularization term to the policy function, namely $\text{log}\pi_\varsigma^{m+1}(a_{m+1}|s_{m+1})$. Then, input $(s_{m+1}, a_{m+1})$ into the two target networks $\varpi_1$ and $\varpi_2$, and obtain the target action value pairs $Q_{\varpi_1}(s_{m+1}, a_{m+1})$ and $Q_{\varpi_2}(s_{m+1}, a_{m+1})$. Take the smaller of the two values, denoted as $Q_{\varpi}(s_{m+1}, a_{m+1})$. 
Finally, updating the critic networks by the SGD algorithm, i.e.,
\begin{equation}
\textcolor{red}{	\begin{split}
		\nabla_{\xi_1}&J_{\xi_1} = \nabla_{\xi_1}Q_{\xi_1}(s_m, a_m) \times \left[ Q_{\xi_1}(s_m, a_m) - r_m \right. \\
		& \left. + \gamma [- \beta \log \pi_\varsigma^{m+1} (a_{m+1} | s_{m+1})+Q_{\varpi}(s_{m+1}, a_{m+1})] \right]
	\end{split}}
	\label{eq36}
\end{equation}
\begin{equation}
\textcolor{red}{	\begin{split}
		\nabla_{\xi_2}&J_{\xi_2} = \nabla_{\xi_2}Q_{\xi_2}(s_m, a_m) \times \left[ Q_{\xi_2}(s_m, a_m) - r_m \right. \\
		& \left. + \gamma [- \beta \log \pi_\varsigma^{m+1} (a_{m+1} | s_{m+1})+Q_{\varpi}(s_{m+1}, a_{m+1})] \right]
	\end{split}}
	\label{eq37}
\end{equation}

We adopt the Adam optimizer to update the actor and critic networks every $I_R$ slots based on $\nabla_{\beta}J_{\beta}$, $\nabla_\varsigma \mathcal{J}_\varsigma$, $\nabla_{\xi_1}J_{\xi_1}$ and $\nabla_{\xi_2}J_{\xi_2}$ in Eqs. \eqref{eq33} - \eqref{eq37}. And now, the process of updating actor and critic networks in slot $t$ are finished and enter to the next slot. Meanwhile, in the process, every $I_t$ slots, we also need to update the parameters of the two target networks $\varpi_1$ and $\varpi_2$ as follows 
\begin{equation}
\textcolor{red}{	\varpi_1 := \omega \xi_1 + (1-\omega)\varpi_1}
	\label{eq38}
\end{equation}
\begin{equation}
\textcolor{red}{	\varpi_2 := \omega \xi_2 + (1-\omega)\varpi_2}
	\label{eq39}
\end{equation}where $\omega \ll 1$ .

Repeat \textbf{\emph{Step 2}} to \textbf{\emph{Step 4}} until episode $e$ reaches to $E^{max}$ and slot $t$ reaches to $S^{max}$, at which point the entire algorithm terminates. Then output a final global model and policy of actor network, denoted as $\varsigma^{\ast}$.

\section{\textcolor{red}{Simulation results}}
\label{sec6}

\subsection{Base settings}
This experiment is entirely based on the scenario described in Section \ref{sec3}, and utilizes Python 3.10 for simulation. The actor network and both critic networks are implemented as four-layer fully connected DNNs, each featuring two hidden layer sand containing 512 neurons. We set up a scenario with four intersections. In urban environments, the variety of object categories is relatively limited, and images of objects within the same category are frequently captured. Therefore, we chose the CIFAR-10 dataset, which contains 10 common categories with 5,000 distinct images per category.

The following training simulation results represent the average of three  experiments, and the environmental parameters and SAC algorithm hyper-parameters used in this paper are shown in Table \ref{tab2}.

\begin{table}[h]\footnotesize
	\centering
	\begin{threeparttable}
		\setlength{\abovecaptionskip}{0pt}
		\caption{Hyper-parameter.}
			\label{tab2}
		\begin{tabular}{p{2cm} p{1.5cm} | p{2cm} p{1.5cm}}
			\hline
			\textbf{Parameter} & \textbf{Value} & \textbf{Parameter} & \textbf{Value} \\
			\hline
			$\vartheta_1/\vartheta_2/\vartheta_3$ & 10/0.001/0.01 & $\psi_1/\psi_2/\psi_3$ & 0.3/0.3/0.4 \\
			$E^{max}$ & 3000 & $S^{max}$ & 100 \\
			$T$ & 1s & $T'$ & 0.005T$\sim$ T \\
			$Z$ & $1500\text{KB}$ & $\mathcal{Z}$ & 512 \\
			$f^{min}$ & $5\times{10}^7\text{Hz}$ & $f^{max}$ & $4\times{10}^8\text{Hz}$ \\
			$p^{min}$ & 5W & $p^{max}$ & 200W \\
			$v^{min}$ & 10m/s & $v^{max}$ & 15m/s \\
			$\tau_1$ & 0.1 & $\tau_2$ & 1 \\
			$\mu$ & 0 & $\sigma$ & 8 \\
			$m$ & 0.023 & $\kappa$ & ${10}^{-27}$ \\
			$\mathcal{B}$ & 256 & $M$ & 64 \\
			$N_0$ & -114dB & $\mathcal{N}$ & 512 \\
			$D$ & 11.2M & $t^{max}$ & 0.02 \\
			$I_R$ & 2 & $I_t$ & 80 \\
			$c$ & 1600 $\text{cyc/s}$ & $R$ & $10^6$ \\
			$\text{Initial learning rate}$ & 0.01 & $\text{momentum of SGD}$ & 0.9 \\
			$f_{t,R}^{comp}$ & $6\times{10}^9\text{Hz}$ & $\gamma$ & 0.99 \\
			$q_0$ & 0.005 & \text{total bandwidth} & $2\times{10}^6\text{Hz}$ \\
			\hline
\end{tabular}
\end{threeparttable}
\end{table}

\subsection{Comparative algorithms}
To conduct comparative experiments, we selected three additional DRL algorithms suitable for continuous action spaces: DDPG, TD3, and Proximal Policy Optimization (PPO).
\begin{itemize}
	\item DDPG: DDPG is a deterministic policy gradient algorithm based on deep learning. It combines the advantages of policy gradient methods and Q-learning, allowing for direct learning of the policy function without the need to explicitly learn the value function.
\end{itemize}

\begin{itemize}
	\item TD3: TD3 is an improvement and extension of DDPG. This algorithm reduces the variance in Q-value estimation by introducing twin Q-networks. Additionally, it improves training stability by incorporating target policy smoothing, which introduces noise when selecting target policy actions.
\end{itemize}

\begin{itemize}
	\item PPO: PPO is a reinforcement learning method based on policy optimization. The core idea of PPO is to directly optimize the policy function without explicitly learning the value function, in order to maximize cumulative rewards. By controlling the step size of policy parameter updates, PPO ensures that the difference between the new and old policies remains moderate, thereby enhancing training stability and convergence speed.
	
\end{itemize}

\subsection{Complexity analysis of comparative algorithms}

\begin{itemize}
	\item SAC: In each update step, the SAC algorithm needs to update the policy of actor network and two critics networks, resulting in a higher time complexity, making the per-step computational complexity approximately $\mathcal{O}\left(3\mathcal{N}\right)$, where $\mathcal{N}$ is the size of the neural network. SAC needs to store one actor network, two critic networks, and two target networks, leading to a space complexity of approximately $\mathcal{O}\left(5\mathcal{N}\right)$. 
	
\end{itemize}

\begin{itemize}
	\item DDPG: In each update step, DDPG needs to update the policy network and one Q-value network, with a time complexity of approximately $\mathcal{O}\left(2\mathcal{N}\right)$. DDPG requires storing one policy network, one target policy network, one Q-value network, and one target Q-value network, resulting in a space complexity of approximately $\mathcal{O}\left(4\mathcal{N}\right)$.
\end{itemize}

\begin{itemize}
	\item TD3: In each update step, TD3 needs to update the policy network and two Q-value networks, with a time complexity of approximately $\mathcal{O}\left(3\mathcal{N}\right)$. TD3 requires storing one policy network, one target policy network, two Q-value networks, and two target Q-value networks, leading to a space complexity of approximately $\mathcal{O}\left(6\mathcal{N}\right)$.
\end{itemize}

\begin{itemize}
	\item PPO: PPO uses mini batch gradient descent and a clipped probability ratio to stabilize training. The computational complexity per step is related to the neural network size $\mathcal{N}$, the number of optimizer iterations $K$, and the mini batch size $M$, resulting in a time complexity of $\mathcal{O}\left(\mathcal{N}KM\right)$. PPO requires storing a policy network and a value network, leading to a space complexity of approximately $\mathcal{O}\left(2\mathcal{N}\right)$.
	
\end{itemize}

\subsection{Simulation results}

\begin{figure}[H]
	\centering
	\begin{subfigure}[b]{\linewidth}
		\centering
		\includegraphics[scale=0.4, trim=25 5 55 40, clip]{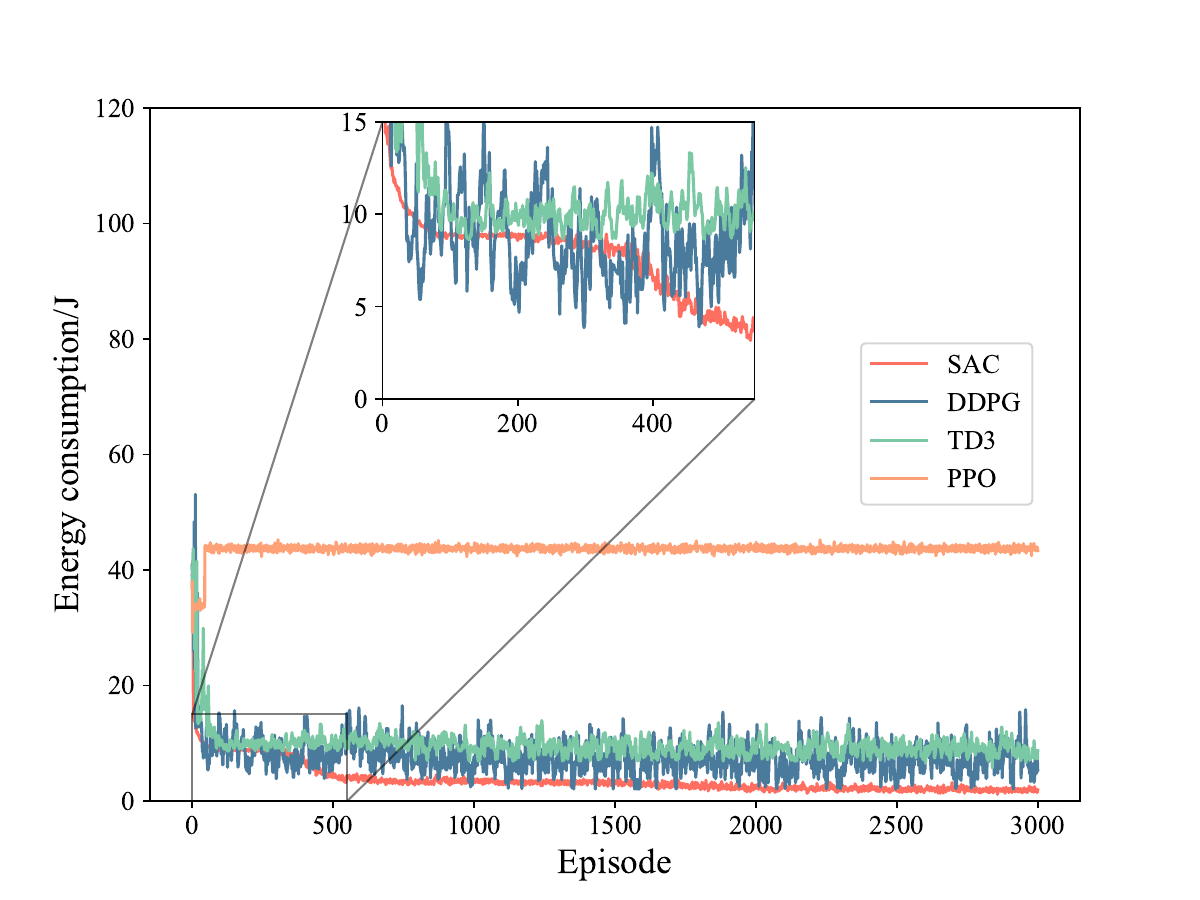}
		\caption{Energy consumption}
		\label{fig3a}
	\end{subfigure}

	\begin{subfigure}[b]{\linewidth}
		\centering
		\includegraphics[scale=0.4, trim=25 5 55 40, clip]{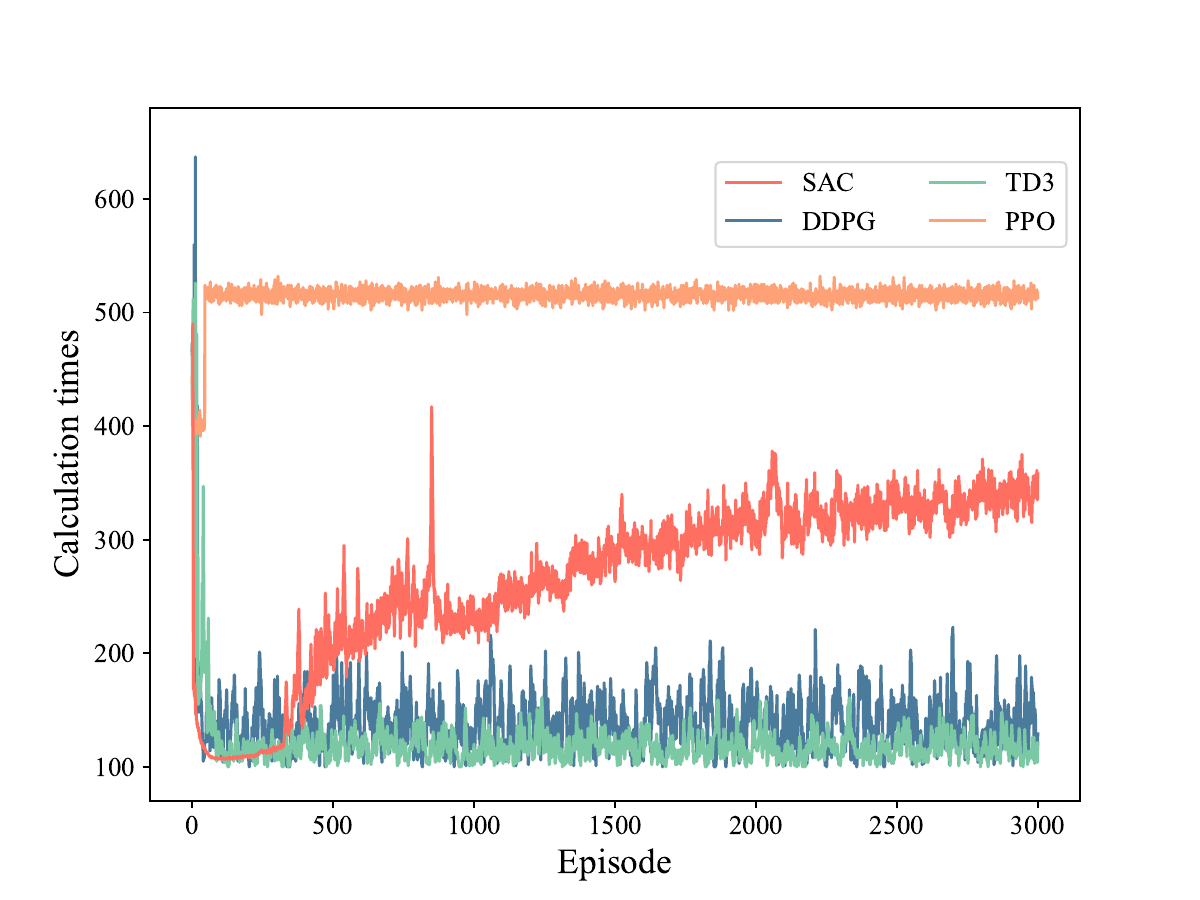}
		\caption{Calculation comparison}
		\label{fig3b}
	\end{subfigure}
	
	\begin{subfigure}[b]{\linewidth}
		\centering
		\includegraphics[scale=0.4, trim=10 5 55 40, clip]{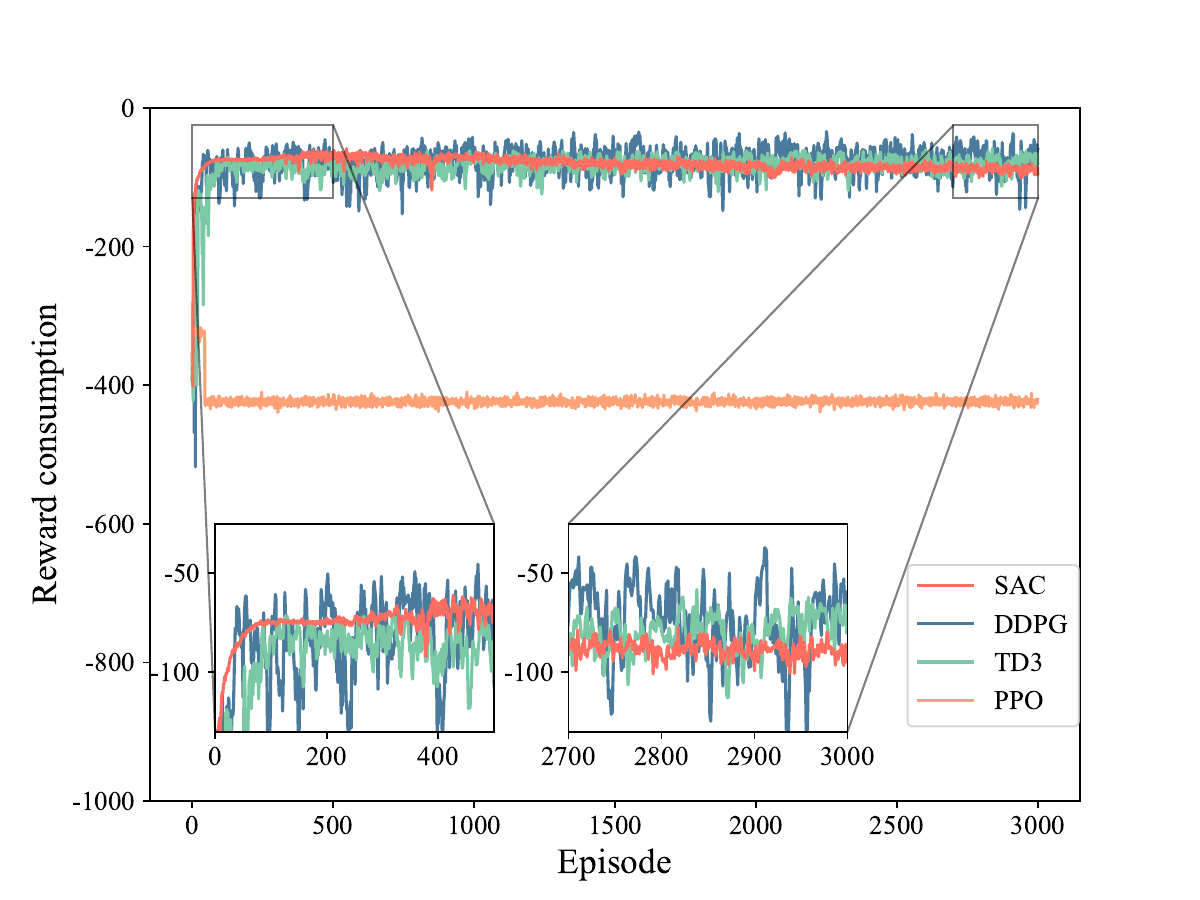}
		\caption{Reward comparison}
		\label{fig3c}
	\end{subfigure}
	\caption{\textcolor{red}{Comparison between different DRL algorithms}}
	\label{fig3}
\end{figure}

\begin{figure}[t]
	\centering
	\begin{subfigure}[b]{\linewidth}
		\centering
		\includegraphics[scale=0.4, trim=25 5 55 40, clip]{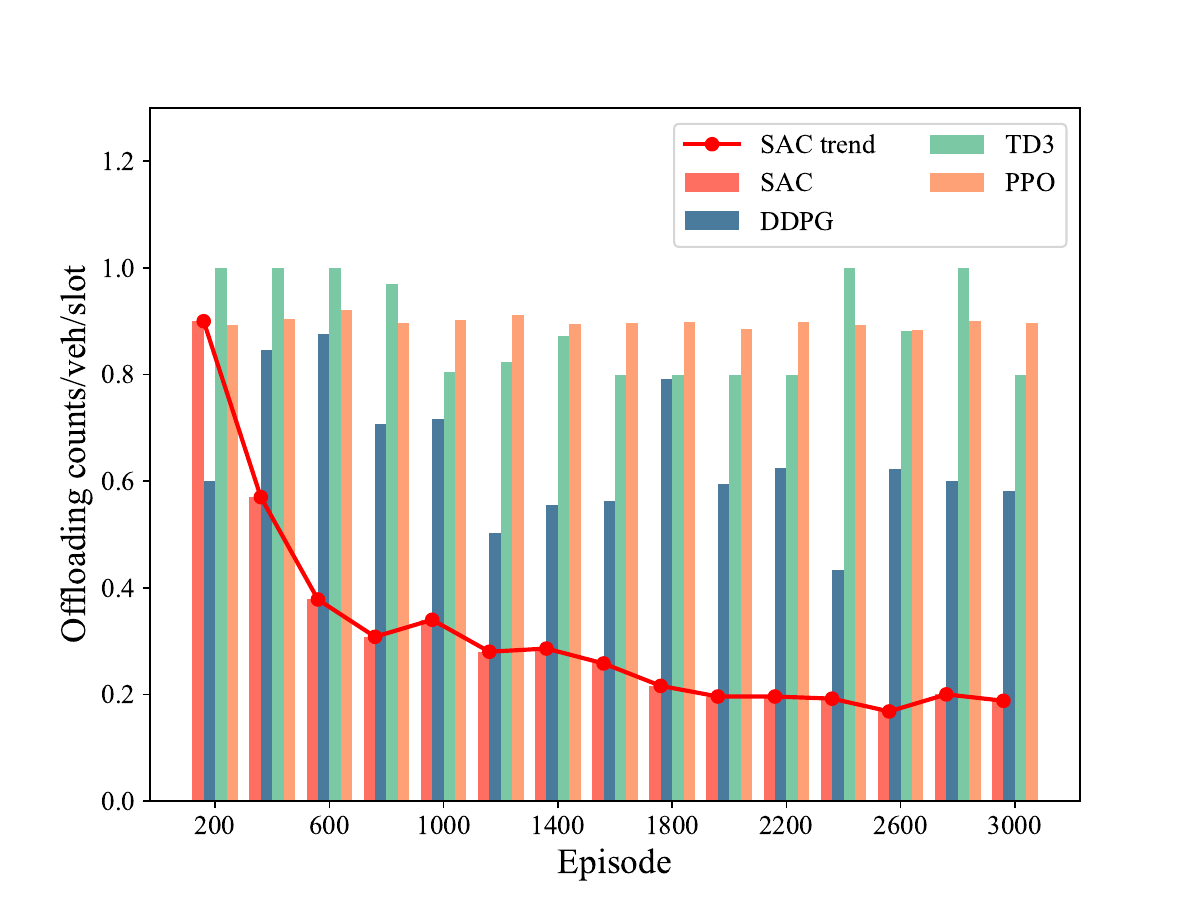}
		\caption{Offloading counts}
		\label{fig4a}
	\end{subfigure}

	\begin{subfigure}[b]{\linewidth}
		\centering
		\includegraphics[scale=0.4, trim=25 5 55 40, clip]{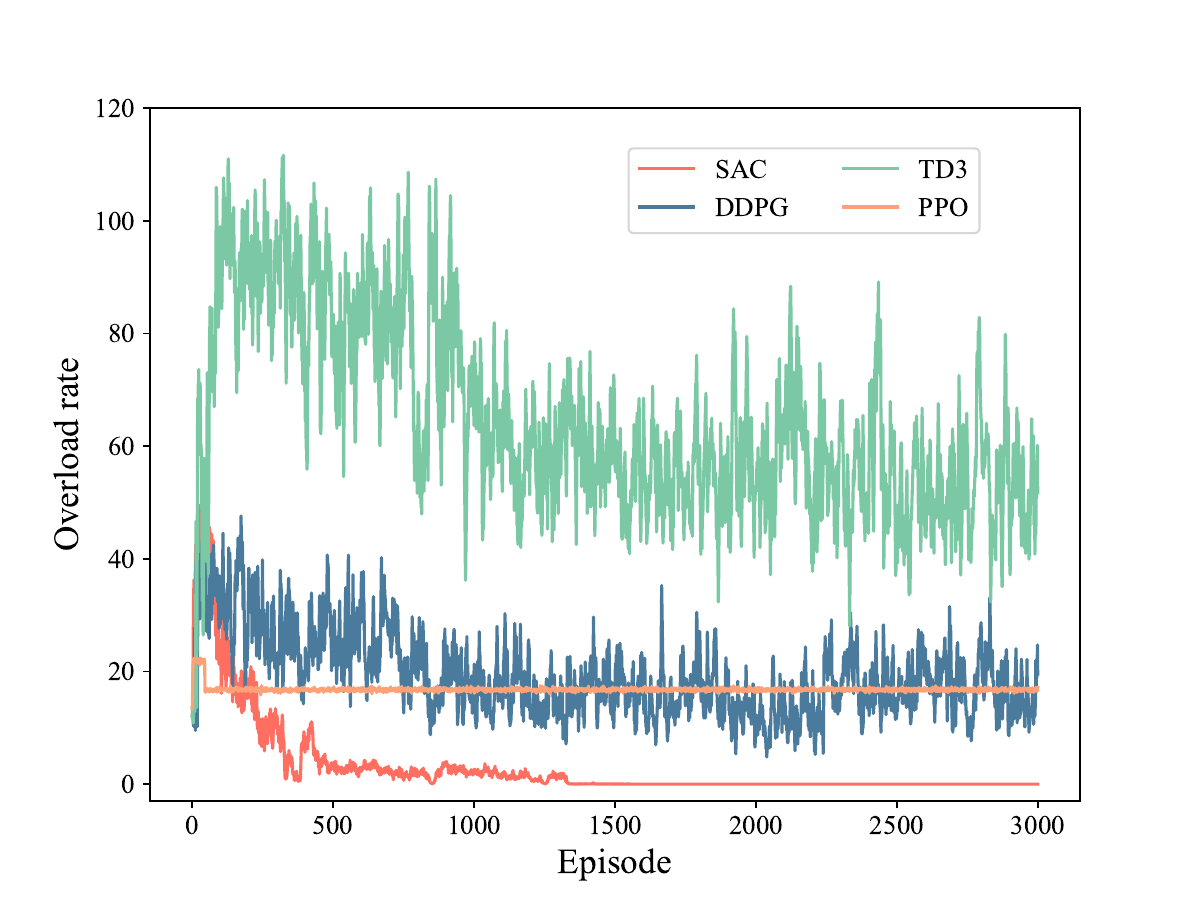}
		\caption{Overload rate}
		\label{fig4b}
	\end{subfigure}
	\caption{\textcolor{red}{Comparison under different evaluation metrics}}
	\label{fig4}
\end{figure}

In Fig. \ref{fig3}, we show a comparison of different DRL methods with five vehicles taking part in each round of FL. The training methods include SAC, DDPG, TD3, and PPO. We assess these methods in terms of energy consumption, computation times, and reward. Regarding energy consumption, as shown in Fig. \ref{fig3a}, PPO algorithm fails to effectively reduce energy consumption, with its energy consumption fluctuating around $(43.69\pm1.24)J$. In contrast, the energy consumption of the SAC, DDPG, and TD3 algorithms shows a rapid decline and gradually converges to a relatively stable value. Specifically, the SAC algorithm exhibits the fastest convergence speed and is more stable compared to other two algorithms, with the final energy consumption stabilizing at $(2.18\pm1.02)J$. The inset provides detailed information for the episodes range $\left[1, 500\right]$. It is evident from the inset that TD3 algorithm exhibits less fluctuation than DDPG algorithms but DDPG algorithm can explore lower energy consumption values. Ultimately, the energy consumption of the DDPG algorithm stabilizes at $(8.95\pm6.87)J$, while that of the TD3 algorithm stabilizes at $(6.24\pm3.54)J$.

As shown in Fig. \ref{fig3b}, in terms of computation times, all four algorithms eventually converged to a stable value. The results show that the computation times of the SAC algorithm gradually stabilized at $338\pm36$ times, significantly higher than other two algorithms. The final computation times of the DDPG and TD3 algorithms are similar, at $162\pm61$ times and $126\pm26$ times, respectively, but the DDPG algorithm exhibits greater fluctuation. From the figure, we can also observe that PPO algorithm has the highest computation times, stabilizing at $516\pm14$ times. Theoretically, more computations result in a more accurate global model. However, combined with the results of energy consumption shown in Fig. \ref{fig3a}, PPO algorithm also has the highest energy consumption. This suggests that PPO is not an ideal choice when the goal is to reduce energy consumption. Additionally, Fig. \ref{fig3c} shows the reward function of these methods. It is clear that all four algorithms eventually converge to a relatively stable reward value. The inset in the lower left corner provides detailed information for the episode range $\left[1, 500\right]$, clearly showing that the reward value of SAC algorithm quickly converged and stabilized after 76 episodes. The DDPG and TD3 algorithms also reach to a relatively stable state after 150 episodes. The inset in the right shows that the final reward values of the SAC, DDPG, and TD3 algorithms stabilized at $-88.07\pm12.79$, $-91.62\pm54.56$, and $-85.53\pm27.36$, respectively. It is also evident from the inset that the reward values of the DDPG and TD3 algorithms tend to surpass that of the SAC algorithm, but in terms of the final value of energy consumption and stability, the SAC algorithm remains the best.

\begin{figure}
	\centering
	\includegraphics[scale=0.4, trim=25 5 55 40, clip]{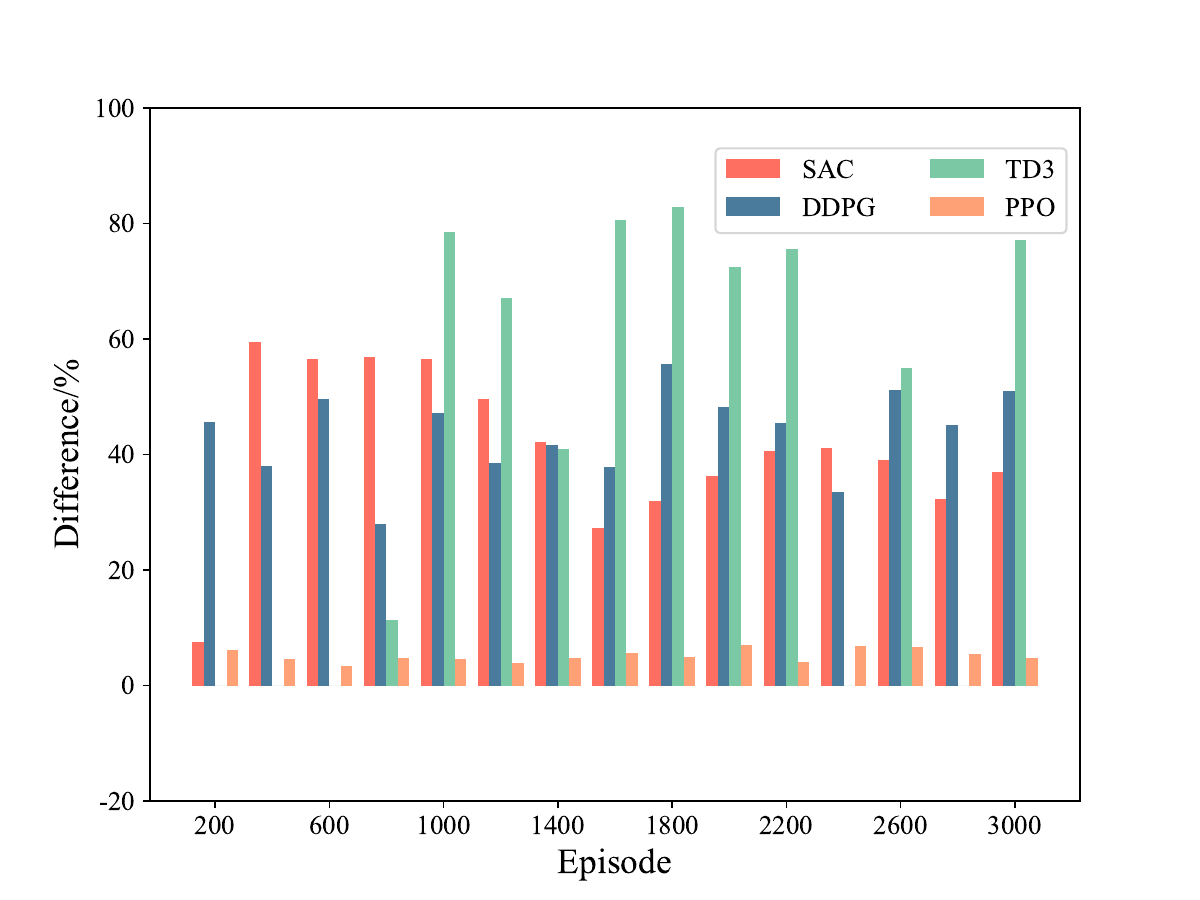}
	\caption{The difference between with and without offloading threshold}
	\label{fig5}
\end{figure}

Overall, the SAC algorithm demonstrates the lowest energy consumption. Although its final reward is slightly lower than those of the DDPG and TD3 algorithms, the SAC algorithm converges more rapidly and stably across all the three considered metrics. This effectiveness is primarily due to SAC's ability to balance exploration and exploitation during policy updates, enhancing convergence efficiency. Consequently, SAC proves to be the most effective algorithm for achieving the goal of minimizing energy consumption, making it the best performer overall.

Fig. \ref{fig4} presents the offloading counts for each vehicle in each slot and overload ratio $\mathcal{R}_0$ under different DRL algorithms. Fig. \ref{fig4a} shows that the offloading counts for the DDPG, TD3, and PPO algorithms fluctuate and do not exhibit a convergence trend, whereas the offloading counts for the SAC algorithm demonstrate a stable decline and eventually converge. Fig. \ref{fig4b} shows that, except for PPO, the overload ratio $\mathcal{R}_0$ of other three algorithms decreases, with the overload ratio of SAC algorithm approaching zero after the 1332 episodes. From the analysis of Fig. \ref{fig4a} and Fig. \ref{fig4b}, it can be seen that the SAC algorithm gradually reduces the offloading counts during training, favoring more local training on the vehicles, thus reducing the overload ratio and eventually achieving a balance between local vehicle training and offloading to the RSU. Although the DDPG and TD3 algorithms do not significantly reduce the offloading counts during training, they can still lower the overload ratio. This indicates that these two algorithms tend to offload training tasks to the RSU for processing during the initial phase of training. Combined with Fig. \ref{fig3a}, this also explains why the energy consumption of these two algorithms is higher than that of the SAC algorithm.
From Fig. \ref{fig4}, we can also observe that PPO algorithm tends to offload tasks to the RSU for processing, but the number of each offloading tasks do not significantly exceed the total tasks in vehicle side. Combined with Fig. \ref{fig3}, we can see that this policy of PPO algorithm increases transmission costs. Therefore, the PPO algorithm consumes more energy.

Fig. \ref{fig5} illustrates the differences between offloading efficiency with and without the threshold $q_0$. From the overall data, it is evident that all values are above $0$, indicating that the offloading efficiency with the threshold $q_0$ is consistently higher than without it. This phenomenon validates that our proposed method of setting the threshold $q_0$ can significantly enhance offloading efficiency.
Specifically, among all algorithms, the TD3 algorithm exhibits the greatest improvement in offloading efficiency, followed by the SAC and DDPG algorithms, with the PPO algorithm showing the least improvement. The superior performance of the TD3 algorithm may stem from its twin delayed deep deterministic policy gradient mechanism, which effectively reduces the variance during the policy network update, thereby enhancing the accuracy of offloading decisions. The SAC algorithm, by introducing the concept of entropy and encouraging exploration diversity, also improves offloading efficiency to a certain extent. In contrast, although the PPO algorithm provides a degree of stability and safety, its relatively conservative policy updates may limit its performance in complex environments.	

\begin{figure}[H]
	\centering
	\begin{subfigure}[b]{\linewidth}
		\centering
		\includegraphics[scale=0.4, trim=25 5 55 50, clip]{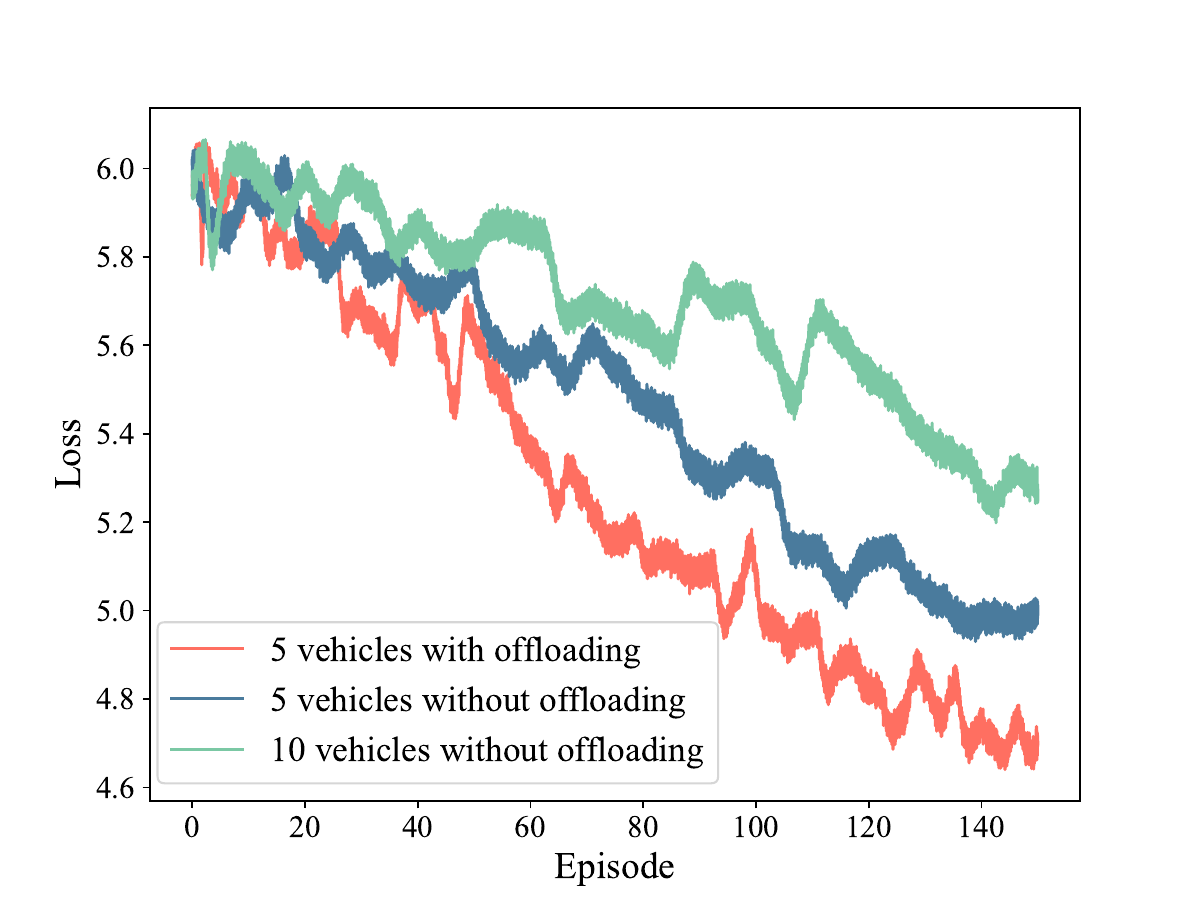}
		\caption{Loss comparison}
		\label{fig9a}
	\end{subfigure}
	
	\begin{subfigure}[b]{\linewidth}
		\centering
		\includegraphics[scale=0.4, trim=25 5 55 50, clip]{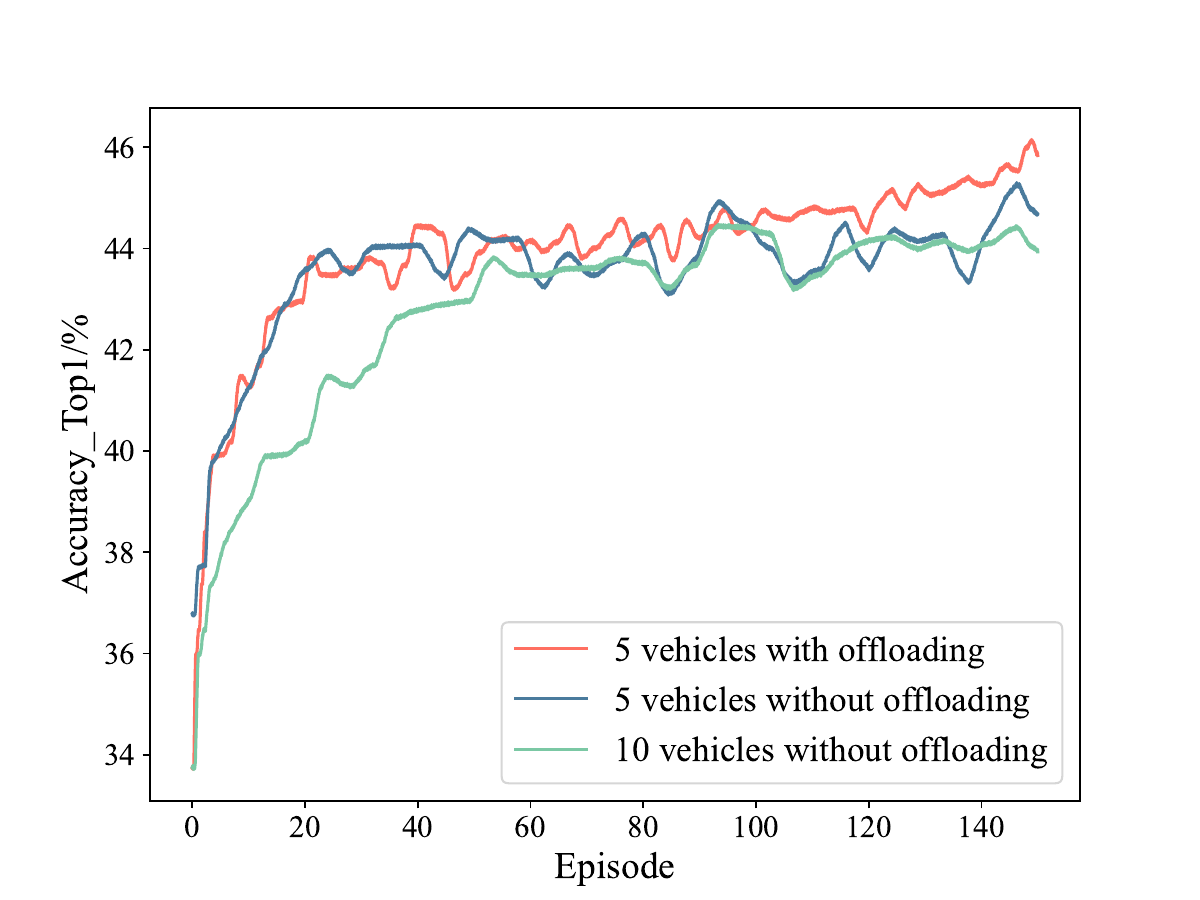}
		\caption{Accuracy of Top1}
		\label{fig9b}
	\end{subfigure}
	
	\begin{subfigure}[b]{\linewidth}
		\centering
		\includegraphics[scale=0.4, trim=25 5 55 50, clip]{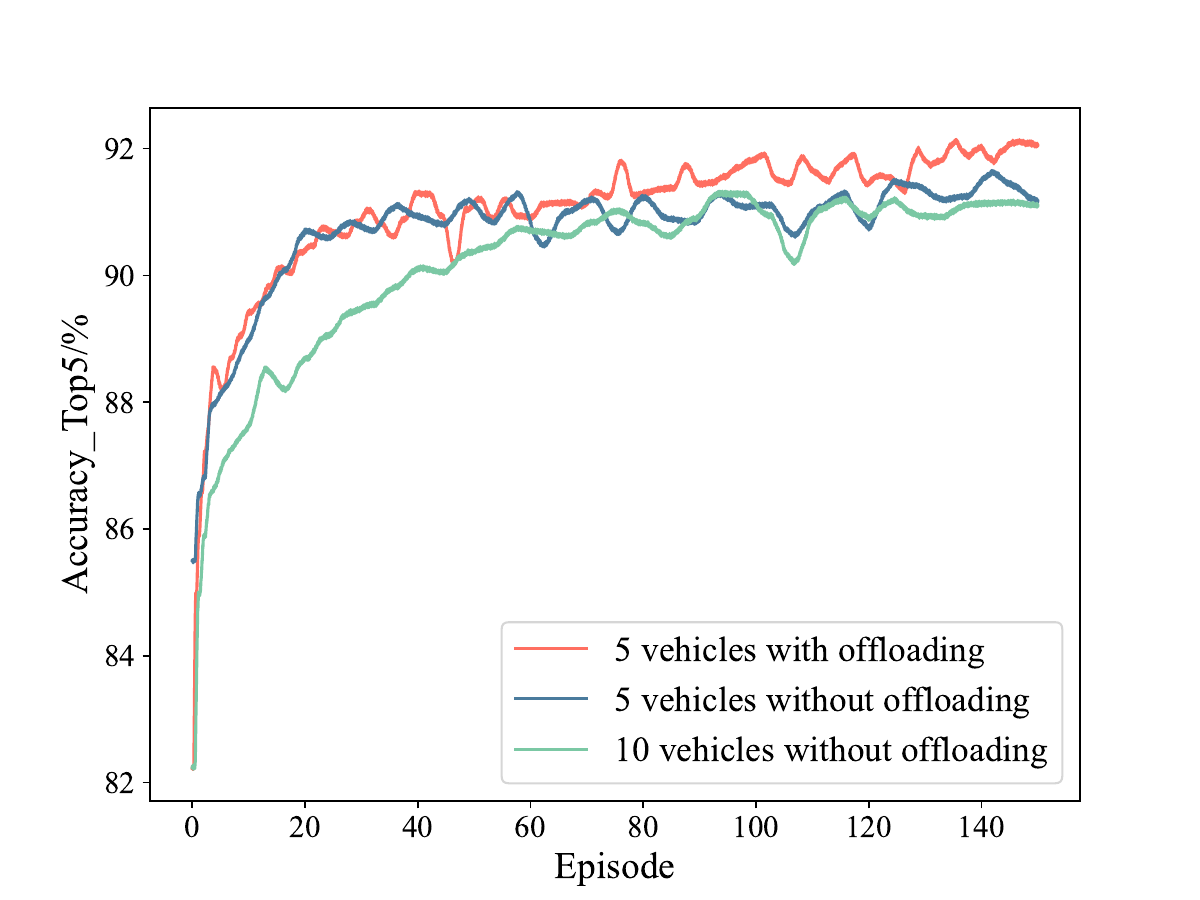}
		\caption{Accuracy of Top5}
		\label{fig9c}
	\end{subfigure}
	\caption{Comparison of loss function and accuracy. In the case with task offloading, when 5 vehicles participate in federated SSL, in each round of training, the model aggregation involves 10 models; whereas, such as FLSimCo, without task offloading, the aggregation involves only 5 models. Similarly, when 10 vehicles are involved in training without task offloading, the model aggregation includes 10 models.}
	\label{fig9}
\end{figure}

\begin{figure}[t]
	\centering
	\begin{subfigure}[b]{\linewidth}
		\centering
		\includegraphics[scale=0.4, trim=3 0 25 50, clip]{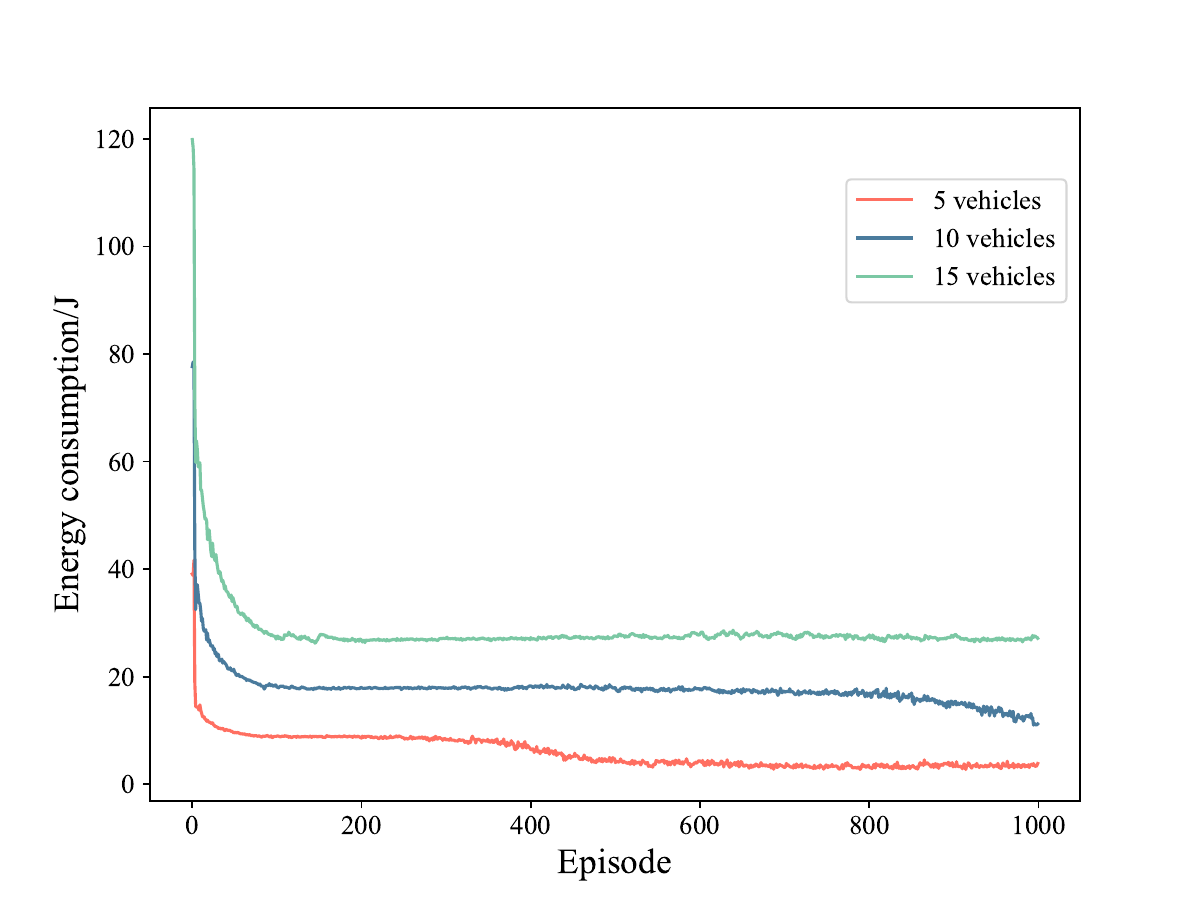}
		\caption{Energy consumption comparison}
		\label{fig6a}
	\end{subfigure}

	\begin{subfigure}[b]{\linewidth}
		\centering
		\includegraphics[scale=0.4, trim=3 0 25 50, clip]{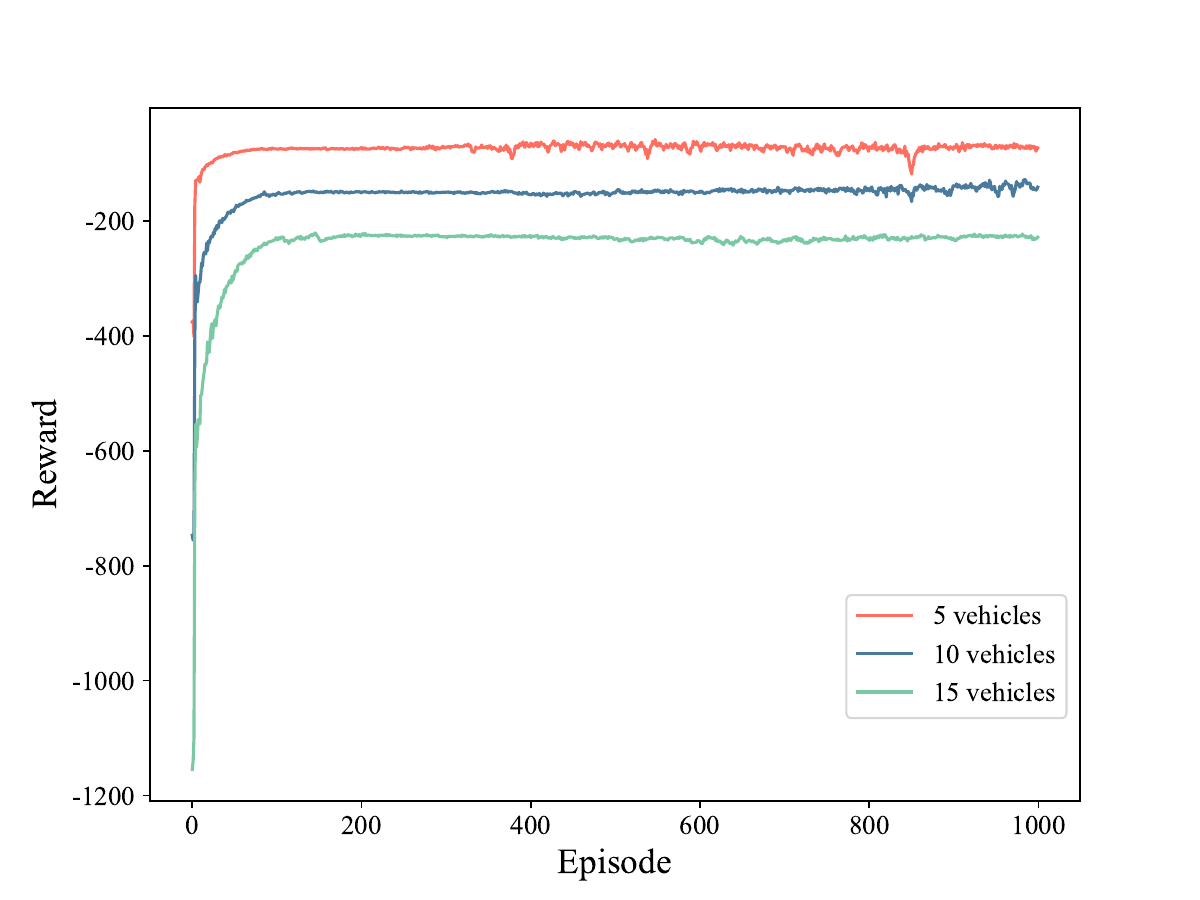}
		\caption{Reward function}
		\label{fig6b}
	\end{subfigure}
	\caption{Comparison between different numbers of vehicles}
	\label{fig6}
\end{figure}

In Fig. \ref{fig9}, we compare the trends in the loss function, the Top1 and Top5 classification accuracy of the global model of FLSimCo (without offloading) and our method (with offloading) under varying numbers of vehicles and local iterations per round. It is obvious that the loss function gradually decreases and converges, while the accuracy steadily increases and converges. Specifically, as illustrated in Fig. \ref{fig9a}, when comparing the cases of 5 vehicles (with 5 models aggregated per round) and 10 vehicles (with 10 models aggregated per round) without task offloading, we observe that with 10 vehicles, the loss function fluctuates more and converges more slowly. This indicates that as the number of vehicles increases, the model diversity also increases, negatively impacting the effectiveness of federated aggregation.
However, by offloading tasks to the RSU for training, we obtain models trained by the RSU, which are then sent back to the BS for aggregation along with the locally trained models. Therefore, with 5 vehicles and task offloading, the final number of models is doubled, resulting in 10 local models aggregated per round, which is the same as the number of models aggregated by the BS with 10 vehicles without task offloading. From the figure, we can see that the 5 vehicles with task offloading (10 models in total) not only outperform the 10 vehicles without task offloading but also the 5 vehicles without task offloading. This is because offloading some tasks to the RSU allows the completion of all scheduled iterations, and more local iterations help mitigate the negative impact of model diversity on federated aggregation.
Fig. \ref{fig9b} and Fig. \ref{fig9c} show that in the first 60 episodes, the fewer models aggregated per round, the faster the accuracy improves. However, as the number of episodes increases, this diversity gradually diminishes, resulting in the accuracy of global with 5 and 10 vehicles without task offloading gradually converging. Additionally, when 5 vehicles participate in federated SSL and offload partial iterations task to RSU, the number of models used in the final aggregation is 10. During the first 60 episodes, the accuracy is not affected by model diversity. Instead, it increases rapidly and steadily. In the later stages of training, due to more iterations per round, the accuracy improvement is significantly better than others. This indicates that a greater number of iterations mitigates the negative effects of model diversity during aggregation.

\begin{figure*}[t]
	\centering 
	\includegraphics[scale=0.4, trim=5 10 5 10, clip]{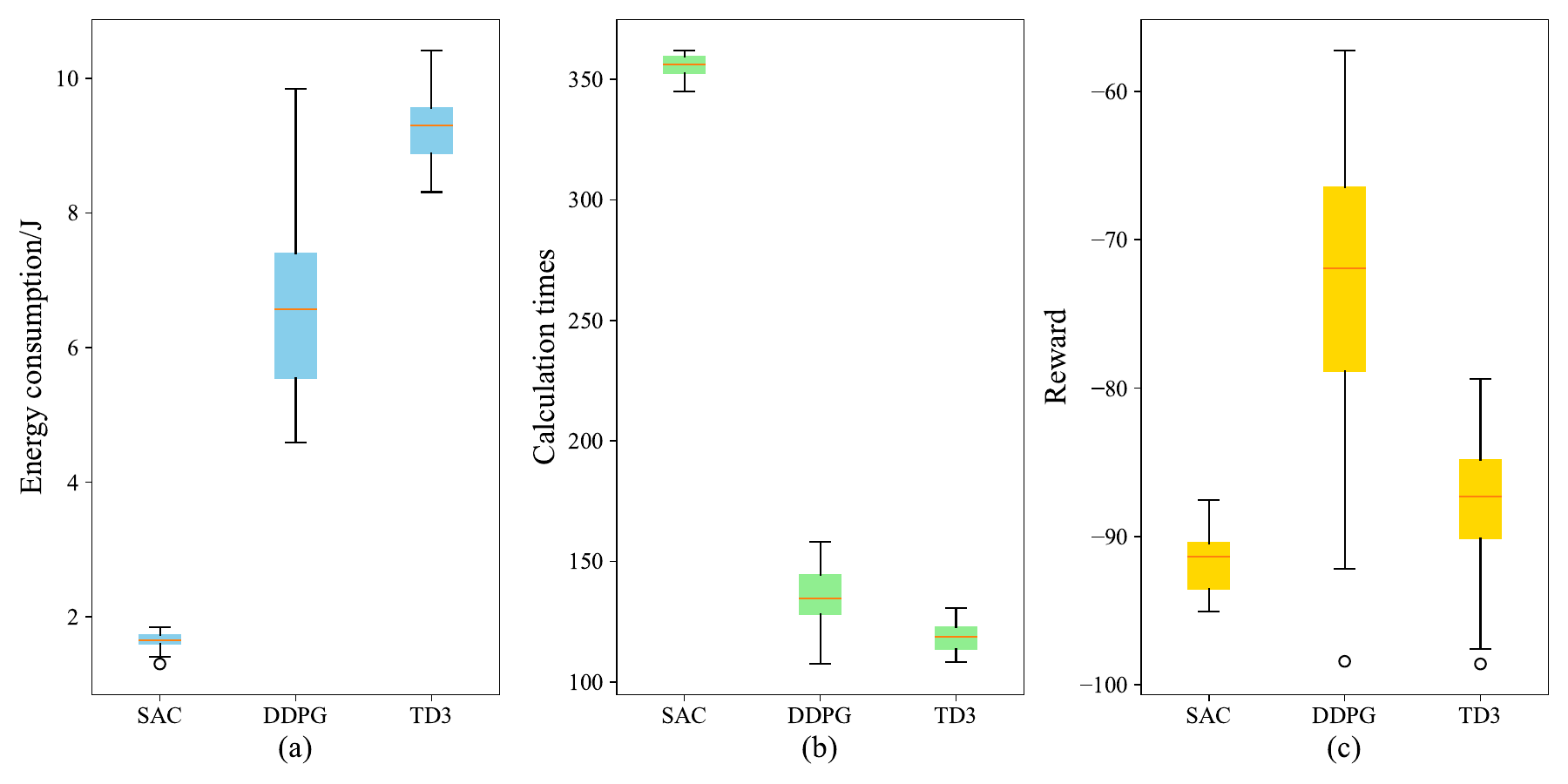}
	\caption{Test results}
	\label{fig7}
\end{figure*}

In Fig. \ref{fig6}, we compare the energy consumption and reward obtained by the SAC algorithm with different numbers of vehicles. From Fig. \ref{fig6a} and Fig. \ref{fig6b}, it can be observed that the energy consumption curve shows a decreasing trend and gradually stabilize and the reward function shows a increasing trend and gradually stabilize. As the number of vehicles increases, the episodes required to reach a stable state also grow, reflecting the heightened complexity of task offloading and resource allocation in multi-vehicle scenarios. Overall, with more vehicles participating in training, total system energy consumption rises, while reward values decline. Additionally, these two figures clearly illustrate the SAC algorithm's superiority in convergence speed and stability. 

We save the models trained by the SAC, DDPG, and TD3 algorithms from Fig. \ref{fig3} and conduct test on them. This test includes four rounds, each consisting of 50 independent experiments. The results of each round are obtained by calculating the mean of these 50 experiments. Finally, we further average the means of these four rounds to derive the overall average results of the comprehensive experiments, which are presented in Fig. \ref{fig7}. In the box plot of Fig. \ref{fig7}, we analyze the results from three aspects: energy consumption, computational times, and reward.
From the box plot in Fig. \ref{fig7}, it can be seen that the test results of the models trained by the SAC algorithm have a lower degree of dispersion, indicating that the data is concentrated and the fluctuation range is small. In contrast, the test results of the models trained by the DDPG algorithm are relatively dispersed, indicating a certain degree of variability. Specifically, in Fig. \ref{fig7}(a), although there is an outlier in the SAC algorithm, this outlier is relatively close to the lower whisker, and the overall data values are concentrated within a small range, showing a compact distribution with few outliers. Therefore, this outlier can be considered negligible for the overall analysis. However, in Fig. \ref{fig7}(c), both the DDPG and TD3 algorithms have outliers, especially the DDPG algorithm, where the outlier is far from the lower whisker, further indicating the relative instability of these two algorithms.
In summary, the models trained by the SAC algorithm demonstrate higher stability and superior performance.

\section{Conclusions}
\label{sec7}

In this paper, we proposed an offloading task and resource allocation algorithm based on the SAC algorithm in federated SSL. It employed this algorithm to allocate the transmission powers from vehicles to BS and RSU, CPU frequency for vehicle local training, assignment ratio of RSU computing resource. Based on the assignment ratio and the computing capability of RSU, this algorithm divided the local iterations task of vehicle into two parts: training locally and offloaded to the RSU. We also aimed to minimize the total energy consumption, including local training, RSU training, and transmission. Additionally, we set an offloading threshold. When the assignment ratio is below this threshold, the vehicle will not perform offloading operation. The conclusions in this paper are as follows
\begin{itemize}
	\item Our proposed new algorithm is suitable for mobile environments and can support real-time decision-making based on the current environmental state.
\end{itemize}	
\begin{itemize}
	\item Vehicles will only offload tasks to the RSU when the assignment ratio exceeds the offloading threshold. This design avoids unnecessary offloading operations, thereby reducing transmission overhead and communication resource waste. By setting a reasonable offloading threshold, we can optimize system performance and improve offloading efficiency
\end{itemize}	
\begin{itemize}
	\item Our proposed algorithm dynamically allocates the number of iterations task between local training and offloading to the RSU, and gradually reduces the offloading frequency during exploration, making full use of the local computing resource. At the same time, it minimizes overload and enhances the utilization of RSU computing resource.
\end{itemize}
\begin{itemize}
	\item Our proposed algorithm increased the number of local iterations while minimizing energy consumption. Although the number of models used during aggregation increases, it still improves classification accuracy compared to algorithm that do not offload task, which compensates for the impact of model diversity on the global model during aggregation.
\end{itemize}

While the approach we suggested has shown promising results in smaller-scale IoV settings, it could encounter challenges related to scalability in larger or more intricate networks. In the future, we will further investigate this aspect to enhance the scalability of the scheme.

%


\bibliographystyle{elsarticle-num}
\balance

\bibliography{egbib}
~~~\\
~~~\\







\end{document}